\documentclass[11pt,a4paper]{article}
\usepackage[hyperref]{acl2018}
\usepackage{times}
\usepackage{latexsym}

\usepackage{url}

\aclfinalcopy % Uncomment this line for the final submission
\def\aclpaperid{1636} %  Enter the acl Paper ID here

\usepackage{xcolor}
\usepackage{amsmath}
\usepackage{amsthm}
\usepackage{multirow}
\usepackage{graphicx}
\usepackage{booktabs}
\usepackage{adjustbox}
\usepackage{svg}
\usepackage{tikz}
\usepackage{amsfonts}
\usepackage{adjustbox}
\usepackage{comment}
\usepackage{gb4e} 
\noautomath 
\newif\ifnotes
\notestrue

\newif\ifdraft
\drafttrue

\newcommand{\cameraready}[1]{\textcolor{black}{#1}}
\newcommand{\cam}[1]{\textcolor{black}{#1}}
\newcommand{\camtwo}[1]{\textcolor{black}{#1}}
\newcommand{\campar}[1]{\textcolor{black}{#1}}

\newcommand{\notehide}[1]{}
\newcommand{\texthide}[1]{}

\newcommand{\notedone}[1]{}

\newcommand{\CNSel}{CN5Sel}
\newcommand{\CNSelected}{CN5Sel(ected)}
\newcommand{\Supplement}{Supplement}

\usepackage{color,soul}
\setul{0.5ex}{0.3ex}

\newcommand{\tc}[2]{\setulcolor{#1}\ul{#2}\setulcolor{black}}
\newcommand{\tcf}[2]{\setulcolor{#1}\ul{\textbf{#2}}\setulcolor{black}}

\newcommand{\tcA}[1]{\tc{blue}{#1}}

\newcommand{\tcC}[1]{\tc{purple}{#1}}
\newcommand{\tcD}[1]{\tc{green}{#1}}
\newcommand{\tcE}[1]{\tc{brown}{#1}}
\newcommand{\tcF}[1]{\tc{cyan}{#1}}
\newcommand{\tcG}[1]{\tc{lime}{#1}}
\newcommand{\tcH}[1]{\tc{darkgray}{#1}}

\newcommand{\tcfA}[1]{\tcf{blue}{#1}}
\newcommand{\tcfB}[1]{\tcf{orange}{#1}}
\newcommand{\tcfC}[1]{\tcf{purple}{#1}}

\newcommand{\tcfE}[1]{\tcf{brown}{#1}}

\usepackage[textsize=tiny, textwidth=2.5cm]{todonotes}
\usepackage{marginnote}
\newcounter{todonumber}
\newcommand{\note}[2][]{{%
 \let\marginpar\marginnote%
 \ifodd\value{todonumber}%
   \reversemarginpar%
 \else%
 \fi%
 \todo[#1]{#2}}%
 \stepcounter{todonumber}%
}

\title{Knowledgeable Reader: Enhancing Cloze-Style Reading Comprehension with External Commonsense Knowledge}

\author{Todor Mihaylov \and Anette Frank \\
Research Training Group AIPHES\\
  Department of Computational Linguistics, Heidelberg University \\
  Heidelberg, Germany \\
  \tt\{mihaylov,frank\}@cl.uni-heidelberg.de \\\\
  }

\date{}

\begin{document}
\maketitle

\begin{abstract}
We introduce a neural reading comprehension model that integrates {\em external commonsense knowledge}, encoded as a key-value memory, in a cloze-style setting. 
Instead of relying only on document-to-question interaction or discrete features as in prior work, our model
attends to relevant external knowledge and combines this knowledge with the context representation before inferring
the answer. 
\campar{This allows the model to attract and imply knowledge from an external knowledge source that is not explicitly stated in the text, but that is relevant for inferring the answer.}
Our model improves results over a very strong baseline on a hard \textit{Common Nouns} dataset, making it a strong competitor of much more complex models.
By including knowledge {\em explicitly}, our model 
can also provide \textit{evidence} about the background knowledge used in the RC process.
\end{abstract}

\section{Introduction}
\label{sec:introduction}
\begin{figure}[t!]
\centering
    \includegraphics[width=0.37\textwidth]{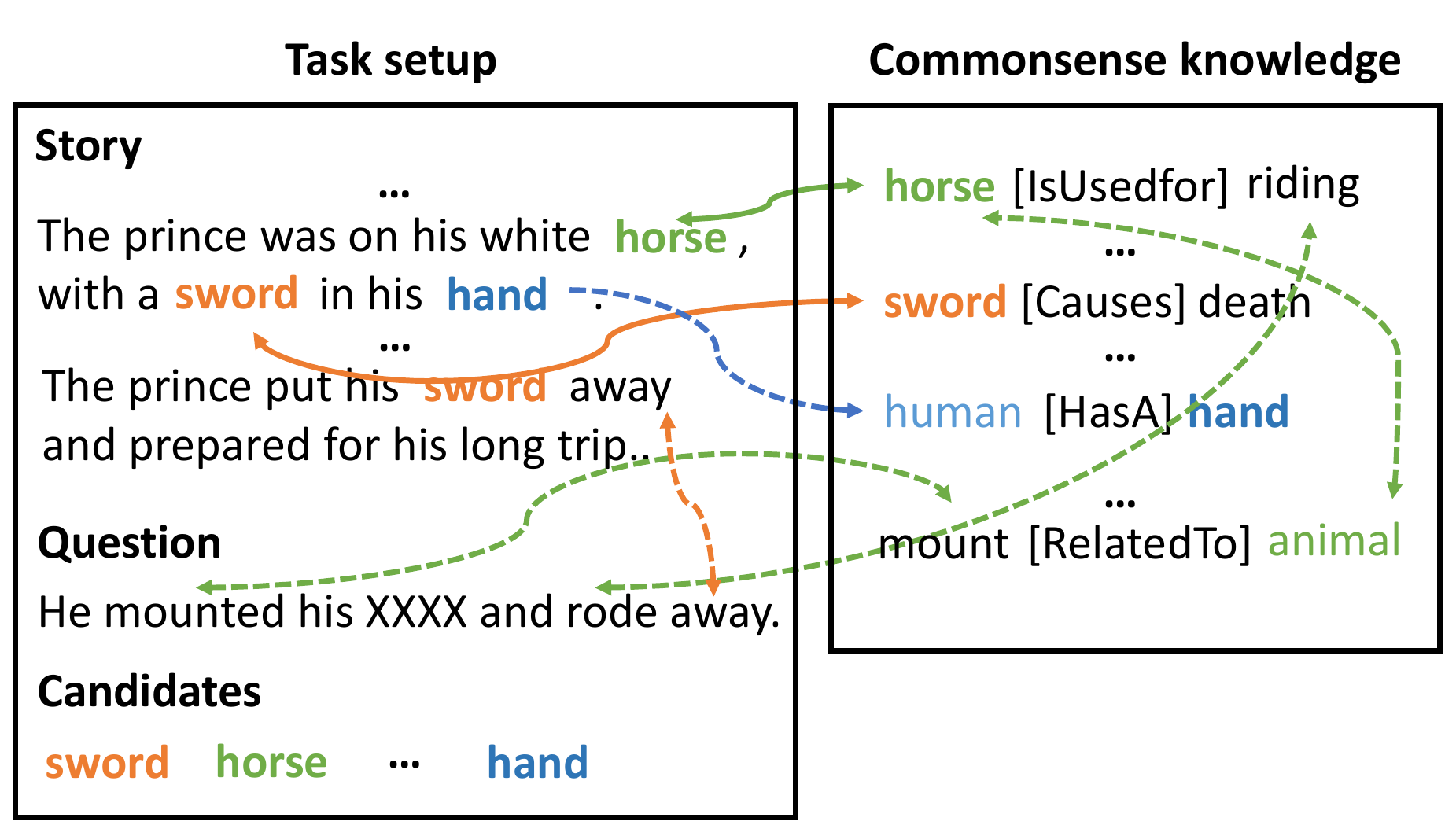}
  
  \caption{Cloze-style reading comprehension with external commonsense knowledge.}
  \label{fig:motivation-example}
\end{figure}

Reading comprehension (RC) is a language understanding task similar to question answering, where a system is expected to read a given passage of text and answer questions about it. Cloze-style reading comprehension is 
a task setting where the question is formed by replacing a token in 
a sentence of the \cameraready{story} with a 
placeholder (left 
\cam{part of} Fig\cam{ure} \ref{fig:motivation-example}). 

In contrast to many previous complex models 
\cite{Weston2015-memorynetworks,Dhingra2016-ga-read,Cui2016-rc-att-over-att,Munkhdalai2016-nse,Sordoni2015}
that perform \emph{multi-turn reading} of a story and a question before inferring the correct answer, we aim to tackle the cloze-style RC task in a way that resembles how humans solve it: using, \cam{in addition,} background knowledge.  
We develop a \cam{neural} model for RC
that can successfully deal with tasks where
most of the information to infer answers from is given in the document \cameraready{(story)}, but where 
additional 
information is needed to 
predict the answer, which can be retrieved from a knowledge base and added to the 
context representations 
explicitly.\footnote{\cam{\emph{`Context representation'} refers to a vector representation computed from textual information only (i.e., document (story) or question).
}}
An 
illustration is given in Fig\cam{ure} \ref{fig:motivation-example}.

Such knowledge may be \emph{commonsense knowledge} or \emph{factual background knowledge about entities and events} that is \cam{not explicitly expressed}
but
can be found in a knowledge base such as ConceptNet \cite{Speer2017Conceptnet55}, BabelNet \cite{NavigliPonzetto:12aij}, Freebase \cite{Tanon2016}
or domain-specific KBs
collected with Information Extraction approaches \cite{Fader2011-IE,Schmitz2012-ie,bhutani-jagadish-radev:2016:EMNLP2016}. 
Thus, we aim to define a neural model that encodes pre-selected knowledge in a memory, and that learns to include the available knowledge as an enrichment to the context representation.

The main difference of our model
to prior state-of-the-art is that instead of relying only on document-to-question interaction  or discrete features 
\cam{while performing}
multiple hops over the document, our model (i) \emph{attends to relevant selected external knowledge} and (ii) \emph{combines this knowledge with the context representation before inferring
the answer}, in a single hop. This allows the model to explicitly imply knowledge that is not stated in the text, but is relevant for inferring the answer, and that can be found in an external knowledge source.
Moreover, by including knowledge explicitly, our model 
\emph{provides evidence and insight about the 
\cam{used} 
knowledge in the RC.}

Our main contributions are:
\textbf{(i)} We develop a method for integrating 
knowledge in a simple but effective reading comprehension model (\textit{AS Reader}, \citet{Kadlec2016-as-reader}) and improve its results significantly whereas other models employ features or multiple hops.
\noindent
\textbf{(ii)} We examine two 
sources of common knowledge: WordNet \cite{WordNet-Miller1990} and ConceptNet \cite{Speer2017Conceptnet55} and show that this type of knowledge is important 
for answering \textit{common noun\cam{s}}
questions and \cam{also} improves slightly the performance for \emph{named entities}.

\textbf{(iii)}
We show \cam{that} 
knowledge facts
can be added directly to the text-only representation, 
enriching the neural context encoding. 
\textbf{(iv)} We demonstrate the effectiveness of the injected knowledge by case studies and data statistics in a qualitative evaluation study.

\section{Reading Comprehension with Background Knowledge Sources}
\label{sec:method}
In this work, we examine the impact of using external knowledge as supporting information 
\cam{for}
the task of \textit{cloze style} reading comprehension.

We build a system with two 
modules. The first, \textit{Knowledge Retrieval}, performs \textit{fact retrieval} and selects a number of facts {\em $f_1$, ..., $f_p$} that might be relevant for connecting story, question and candidate answers. The second, main module, \textit{the Knowledgeable Reader}, is a knowledge-enhanced neural module. It uses the input of the story context tokens \textit{$d_{1..m}$}, the question tokens $q_{1..n}$, the set of answer candidates $a_{1..k}$ and a set of `relevant' background knowledge facts $f_{1..p}$ in order to select the right answer. To include 
external
knowledge for the RC task, we encode each fact $f_{1..p}$ and use attention to select the most relevant among them for each token in the story \cam{and question}. 
We expect that enriching the text with additional knowledge about the 
\cam{mentioned} concepts will improve the prediction of correct answers in a strong \emph{single-pass}
system. 
See Fig\cameraready{ure} \ref{fig:motivation-example} for illustration. 

\subsection{Knowledge Retrieval}
\label{sec:method:know-retrieval}
In our experiments we use knowledge from the Open Mind Common Sense (OMCS, \citet{Singh2002-common-sense-kw-omcs}) part of ConceptNet, a crowd-sourced resource of
commonsense \cam{knowledge}  
with a total of $\sim$630k facts.
Each fact $f_i$ is represented as a triple \textit{$f_i$=(subject, relation, object)}, where \textit{subject} and \emph{object} can be multi-word expressions and \emph{relation} is a relation type. An example is: \textit{([\cam{bow}]$_{subj}$, [IsUsedFor]$_{rel}$, [\cam{hunt, animals}]$_{obj}$)}

We experiment with \cam{three} set-ups:
\cam{using (i) all facts from OMCS that pertain to ConceptNet, referred to as \textit{CN5All}}, 
\cam{(ii) using all facts from \textit{CN5All} excluding some 
WordNet relations
referred to as 
\textit{\CNSel(ected)}} \cam{(see Section \ref{sec:data})},
and using \cam{(iii)} facts from \cam{OMCS} that have  \textit{source} set to  \textit{WordNet} 
(\cam{\textit{CN5WN3})}.

\paragraph{Retrieving relevant knowledge.} 
For each instance ($D$, $Q$, $A_{1..10}$) we retrieve relevant commonsense background facts. 
We first retrieve facts that contain 
lemmas that can be looked up via tokens contained in
any $D$(ocument), $Q$(uestion) or $A$(nswer candidates).
We add a weight value for each node: $4$, if it contains a lemma of a \cam{candidate} token from $A$;
$3$, if it contains a lemma from the tokens of $Q$; and $2$ if it contains a lemma from the tokens of $D$. The selected weights are chosen heuristically such that they model \cam{relative} fact importance 
\cam{in different interactions
} as \textit{A+A $>$ A+Q $>$ A+D $>$ D+Q $>$ D+D}. 
We weight the fact triples that contain these lemmas as nodes, by summing the weights of the subject and object \cam{arguments}. 
Next, we sort the knowledge triples by this overall weight value. 
To limit the memory of our model, we run experiments with different size\cam{s} of the top number of facts ($P$)
selected from all instance \cameraready{fact} candidates, $P \in \{50, 100, 200\}$.
As additional retrieval limitation, we force the number of facts per answer candidate to be the same, in order to 
avoid
a frequency bias for an answer candidate that appears more often in the knowledge source. Thus, if we select the maximum 100 facts for each task instance and we have 10 answer candidates $a_{i=1..10}$, we retrieve the top 10 facts for each candidate $a_i$ that has either a subject or an object lemma 
for a token in $a_i$. If \cam{the} same fact contains
lemmas of two candidates $a_i$ and $a_j$ ($j>i$), we add the fact once for $a_i$ and do not add the same fact again for $a_j$. If several facts have the same weight, we take the first in the order 
of the list\footnote{\cameraready{We also experimented with re-ranking the facts with the same weight sums using tf-idf
but we did not notice a difference in 
performance.}}, i.e., the order of retrieval from the database. If one candidate has less than $10$ \cameraready{facts}, the overall fact candidates for the sample will be less than the maximum (100).

\subsection{Neural Model: Extending the Attention Sum Reader with a Knowledge Memory}
\label{sec:method:neural-model}
We implement our
{\em Knowledgeable Reader (KnReader)} 
using as a bas\camtwo{is}
the \textit{Attention Sum Reader}  as one of the strongest core models for single-hop RC.
We extend it
with a knowledge fact memory that is filled with pre-selected facts. Our aim is to examine how adding 
\cam{commonsense} knowledge to a simple yet effective model can 
improve the RC process
and to show some evidence of that 
by attending on the incorporated knowledge facts. 
The model architecture is shown in Figure \ref{figure:knowledge-able-reader}.

\begin{figure}[t!]
  \centering
  \includegraphics[width=0.50\textwidth]{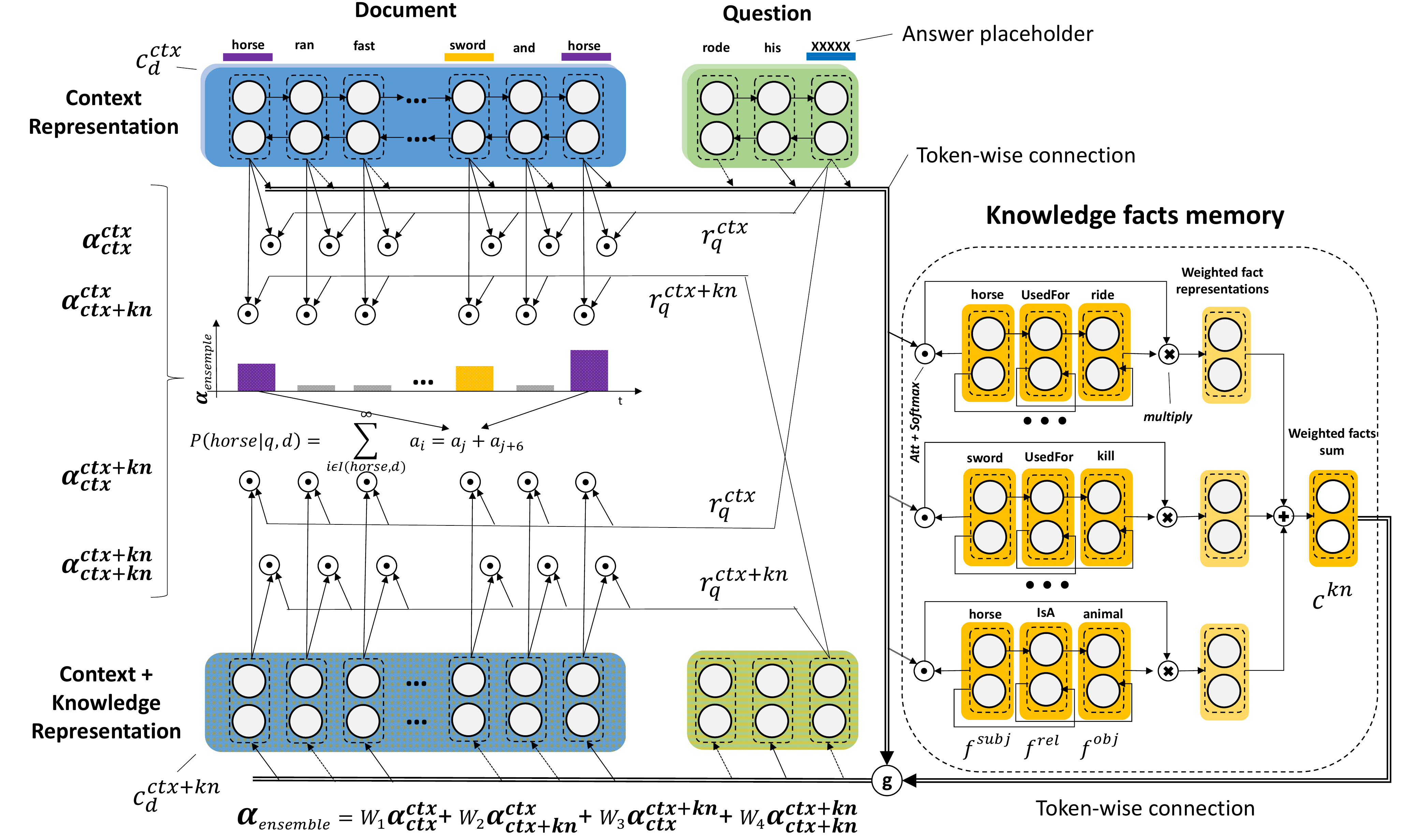}
  \caption{The Knowledgeable Reader combines plain \textit{context} \& \textit{enhanced} (\textit{context + knowledge}) repres.\ of 
  \textit{D} and \textit{Q}
 \cam{and} 
  retrieved knowledge from \cam{the} explicit memory with \cam{the} \textit{Key-Value} approach.}
  \label{figure:knowledge-able-reader}
\end{figure}

\textbf{Base Attention Model.} 
The Attention-Sum Reader \cite{Kadlec2016-as-reader}, our base 
model for RC reads the input of story tokens \textit{$d_{1..n}$}, the question tokens \textit{$q_{1..m}$}, and the set of candidates \textit{$a_{1..10}$} \cam{that occur in the story text.}
The model calculates the attention between the question representation $r_q$ and the story token context encodings of the candidate tokens \textit{$a_{1..10}$} and sums the attention \cam{scores} for the candidates that appear multiple times in the story.
The model 
selects \cam{as answer} 
the candidate that has \cam{the highest attention score}.

\textbf{Word Embeddings Layer.} 
We represent input document and question tokens $w$ by looking up their embedding representations $e_{i} = Emb(w_i)$, where $Emb$ is an embedding lookup function.
We apply dropout \cite{Dropout-Srivastava2014} with keep probability $p=0.8$ to the output of the embeddings \cam{lookup} layer. 

\textbf{Context Representations.} 
To represent the document and question contexts, we first encode the tokens with a Bi-directional GRU (Gated Recurrent Unit) \cite{chung2014-gru} to 
obtain context\cam{-encoded} representations \cam{for document ($c^{ctx}_{d_1..n}$) and question ($c^{ctx}_{q_1..m}$) encoding}: 

\begin{align}
c^{ctx}_{d_1..n} &= BiGRU^{ctx}(e_{d_1..n}) \in \mathbb{R}^{n \times 2h}\\
c^{ctx}_{q_1..m} &= BiGRU^{ctx}(e_{q_1..m}) \in \mathbb{R}^{m \times 2h}
\end{align}
, where $d_i$ and $q_i$ \cam{denote} the $i$th token of a text sequence $d$ (document) and $q$ (question), respectively, $n$ and $m$ is the size of $d$ and $q$ and $h$ the output hidden size \cam(256) of a single $GRU$ unit. 
$BiGRU$ is defined in (\ref{eq:3}), with $e_i$ a word embedding vector 

\begin{align}\label{eq:3}
BiGRU^{ctx}(e_{i}, h_{i_{prev}}) = 
\begin{split}
[\overrightarrow{GRU}(e_{i}, \cam{\overrightarrow{h_{i_{prev}}}}
),\\
\overleftarrow{GRU}(e_{i}, 
\cam{\overleftarrow{h_{i_{prev}}}}
)]
\end{split}
\end{align},
where $h_{i_{prev}} = [\overrightarrow{h_{i_{prev}}}, \overleftarrow{h_{i_{prev}}}]$, and $\overrightarrow{h_{i_{prev}}}$ and $\overleftarrow{h_{i_{prev}}}$ are the previous hidden states of the forward and backward layers. 
Below we use $BiGRU^{ctx}(e_{i})$ without the hidden state\camtwo{, for short}.

\textbf{Question \cam{Query} Representation.~} 
For the question we construct a single 
vector representation 
\cam{$r^{ctx}_{q}$ by retrieving the 
token representation at the placeholder (XXXX) index \textit{pl} (cf.\
Figure \ref{figure:knowledge-able-reader}):} 
\begin{equation}
r^{ctx}_{q} = c^{ctx}_{q_{i..m}}[pl] \in \mathbb{R}^{2h}
\end{equation}
where $[pl]$ is an element pickup operation.

Our question vector representation is different from the original \textit{AS Reader} that builds the question by concatenating the \emph{last states} of a forward and backward layer $[\overrightarrow{GRU}(e_{m}), \overleftarrow{GRU}(e_{1})]$. 
\cameraready{We changed the original representation as we observed some very long questions and in this way aim to prevent the context encoder from 'forgetting' where the placeholder is.}

\textbf{Answer \cam{Prediction}: $Q^{\cam{ctx}}$ to $D^{\cam{ctx}}$ Attention.} 
In order to \cam{predict} the correct answer to the given question, we rank the given answer candidates $a_{1}..a_{L}$ according to the normalized attention sum \cam{score} between the \cam{context ($ctx$) representation} of the question \cam{placeholder} $r^{ctx}_{q}$ and \cam{the representation of} the candidate tokens
\cam{in the document}:
\cam{
\begin{align}
P(a_{i}|q,d) &= \cam{softmax}(\sum \alpha_{i_j}) \label{eq:attsum}\\
\alpha_{i_j} &= \cam{Att(}\cam{r^{ctx}_{q}}, c^{ctx}_{d_j}), i \in [1..L]
\end{align}}
, where $j$ is an index pointer from the list of indices 
that point to the candidate $a_i$ token occurrence\cameraready{s} in the document context representation $c_{d}$. \cam{$Att$} is a dot product. 

\paragraph{Enriching Context Representations with Knowledge (Context+Knowledge).} 
To \cam{enhance} the representation of the context, we add knowledge, retrieved as a set of knowledge facts.

\subparagraph{Knowledge Encoding.~} 
 For each instance in the dataset, we retrieve a number of relevant facts (cf.\
 Section \ref{sec:method:know-retrieval}). Each retrieved fact is represented as a triple $f = (w^{subj}_{1..L_{subj}}, w^{rel}_0, w^{obj}_{1..L_{obj}})$, where $w^{subj}_{1..L_{subj}}$ and $w^{obj}_{1..L_{obj}}$ are a multi-word expressions representing the $subject$ and $object$ with 
 sequence lengths $L_{subj}$ and $L_{obj}$, and $w^{rel}_0$ is a word token corresponding to a relation.\footnote{The $0$ in $w^{rel}_0$ indicates that we encode the relation as a single \textit{relation type} word. Ex. \textit{/r/IsUsedFor}.}
 As a result of fact encoding, we obtain a separate knowledge memory for each instance in the data. 

To encode the knowledge we use 
a $BiGRU$ to encode the 
\cam{triple argument}
\cam{tokens}
into the following 
context\cam{-encoded} representations:
\begin{align}
f^{subj}_{last} = BiGRU(Emb(w^{subj}_{1..L_{subj}}), 0) \\
f^{rel}_{last} = BiGRU(Emb(w^{rel}_{0}), f^{subj}_{last})\\
f^{obj}_{last} = BiGRU(Emb(w^{obj}_{1..L_{subj}}), f^{rel}_{last})
\end{align}

, where $f^{subj}_{last}$, $f^{rel}_{last}$, $f^{obj}_{last}$ are the final hidden 
\cam{states}
of the context encoder $BiGRU$, that are also used as initial representations for the encoding of the next triple attribute in left-to-right order. 
See \textit{\Supplement} for comprehensive visualizations.
The motivation behind this encoding is: 
(i) 
\cam{We e}ncode the knowledge fact attributes in the same vector space as the plain tokens; (ii) 
\cam{we p}reserve the triple directionality; (iii)
we use the relation type as a way of filtering the \textit{subject} information to initialize the \textit{object}. 

\subparagraph{Querying the Knowledge Memory.}
To enrich the context representation of the document and question tokens with the facts collected in the knowledge memory, we select a single \textit{sum of weighted fact representations} for each token using Key-Value retrieval \cite{Miller2016-kv-memnet}. 
In our 
\camtwo{model}  
the \textit{key} $M^{k(ey)}_i$ can be either $f^{subj}_{last}$ or $f^{obj}_{last}$ and the \textit{value} $M^{v(alue)}_i$ is $f^{obj}_{last}$.

For each context-encoded token $c^{ctx}_{s_{i}}$ \cam{($s = {d, q}$;
$i$ 
the token index)} we attend over all knowledge memory keys $M^k_i$ in the retrieved 
\cam{$P$ 
knowledge 
facts}. We use an attention function $Att$, scale the scalar attention value using 
$softmax$, 
multiply it with the value representation $M^v_i$ and sum the result into a single vector value representation \cam{$c^{kn}_{s_{i}}$}:
\cam{
\begin{equation}
c^{kn}_{s_{i}} = \sum softmax(Att(c^{ctx}, M^k_{1..P}))^{T}M^v_{1..P}
\end{equation}} 
\cam{$Att$} is a dot product, but it can be replaced with \cam{an}other attention function. 
As a result of this operation, the context token representation \cam{$c^{ctx}_{s_{i}}$} and the corresponding retrieved knowledge \cam{$c^{kn}_{s_{i}}$} are in the same 
vector
space $\in \mathbb{R}^{2h}$.

\subparagraph{\cam{Combine} Context and Knowledge \cam{$(ctx+kn)$}.}
We combine the
original context token representation $c^{ctx}_{s_{i}}$, 
with the acquired knowledge representation $c^{kn}_{s_{i}}$ to obtain $c^{ctx+kn}_{s_{i}}$:
\begin{equation}
  c^{ctx+kn}_{s_{i}} = \gamma c^{ctx}_{s_{i}}+(1-\gamma)c^{kn}_{s_{i}}
\end{equation},
where \cameraready{$\gamma=0.5$}. \cameraready{We keep $\gamma$ static but it can be replaced with a gating function.}

\subparagraph{Answer \cam{Prediction}: \cam{$Q^{ctx(+kn)}$ to $D^{ctx(+kn)}$}.}
\cam{To rank answer candidates \cam{$a_{1}..a_{L}$} we use attention sum similar to Eq.\ref{eq:attsum} over an attention $\alpha^{ensemble}_{i_j}$ that combines attentions between context ($ctx$) and context+knowledge ($ctx+kn$) 
representations of the question ($r^{ctx(+kn)}_q$) and candidate token occurrences $a_{i_j}$ in the document $c^{ctx(+kn)}_{d_j}$}:
\cam{
\begin{align}
P(a_{i}|q,d) &= softmax(\sum \alpha^{ensemble}_{i_j})\\
\alpha^{ensemble}_{i_j} = \label{eq:ensemble-att} 
\begin{split}
W_{1}Att(r^{ctx}_{q}, c^{ctx}_{d_j})\\
+W_{2}Att(r^{ctx}_{q}, c^{ctx+kn}_{d_j})\\
+W_{3}Att(r^{ctx+kn}_{q}, c^{ctx}_{d_j})\\
+W_{4}Att(r^{ctx+kn}_{q}, c^{ctx+kn}_{d_j})
\end{split}
\end{align}}
\cam{, where $j$ is an index pointer from the list of indices 
that point to the candidate $a_i$ token occurrence\cameraready{s} in the document context representation $c^{ctx(+kn)}_{d}$. 
$W_{1..4}$ are scalar weights \cameraready{initialized with $1.0$ and optimized during training.}\footnote{\cameraready{An example for learned $W_{1..4}$ is 
($2.13$,  %(Dctx, Qctx) 
$1.41$, %(Dctx, Qctx+kn)  
$1.49$, %(Dctx+kn, Qctx)   
$1.84$) % %(Dctx+kn, Qctx+kn) 
%.
\cam{in setting}
(CBT CN, \CNSel, Subj-Obj as k-v, 50 facts).}}
}
We propose the combination of $ctx$ and $ctx+kn$ attentions because our task 
does not provide supervision whether the knowledge is needed or not.

\section{Data and Task Description}
\label{sec:data}

We experiment with knowledge-enhanced cloze-style reading comprehension using the \textit{Common Nouns} and \textit{Named Entities}
partitions
of the Children's Book Test (CBT) dataset \cite{Hill2016-booktest}.

In the CBT cloze-style task a system is asked to read a children story context of 20 sentences. The following 21$^{st}$ sentence involves a placeholder token that the system needs to predict, by choosing from a given set of 10 candidate words from the document. An example with suggested external 
knowledge facts
is given in Figure \ref{fig:motivation-example}. 
While in its \textit{Common Noun\cam{s}} setup, the task can be considered as a language modeling task, 
\citet{Hill2016-booktest} show that humans can answer the questions without the full context with an accuracy of only \textit{64.4\%} and a language model alone with \textit{57.7\%}. By contrast, the human performance when given the full context is at \textit{81.6\%}. Since the best neural model \cite{Munkhdalai2016-nse} achieves only \textit{72.0\%} on the task, we hypothesize that the task itself can benefit from external knowledge.
The characteristics of the data are shown in Table \ref{table:data-cbt}.
\begin{table}[t]
\centering
\begin{tabular}{rcc}
               & \textbf{CN} & \textbf{NE} \\ \hline
\textbf{Train} & 120,769 / 470           & 108,719 / 433             \\
\textbf{Dev}   & 2,000 / 448             & 2,000 / 412               \\
\textbf{Test}  & 2,500 / 461             & 2,500 / 424               \\
\textbf{Vocab} & 53,185                & 53,063                  \\
% \textbf{Human} & 81.6 &    81.6                \\ 
 \hline
\end{tabular}
\caption{Characteristics of Children Book Test datasets. CN: \textit{Common Nouns}, NE: \textit{Named Entities}. Cells for \textit{Train, \cam{Dev, Test}}
%Test, Dev} 
show overall numbers of examples and average story size in tokens.}
\label{table:data-cbt}
\vspace{-2mm}
\end{table}

% \begin{table}[]
% \centering
% \begin{tabular}{lll}
%                & \textbf{Common Nouns} & \textbf{Named Entities} \\ \hline
% \textbf{Train/avg size} & 120,769/470           & 108,719/433             \\
% \textbf{Dev}   & 2,000/448             & 2,000/412               \\
% \textbf{Test}  & 2,500/461             & 2,500/424               \\
% \textbf{Vocab} & 53,185                & 53,063                  \\
% \textbf{Human performance} & 81.6                  & 81.6                    \\ \hline
% \end{tabular}
% \caption{Characteristics of the Children Book Test dataset.}
% \label{table:data-cbt}
% \end{table}

Other
popular cloze-style datasets such as CNN/Daily Mail \cite{Hermann2015-rc-cnn-dm} or WhoDidWhat \cite{Onishi2016-rc-whodidwhat}
are mainly fo\-cu\-sed on finding \textit{Named Entities} where the benefit of adding commonsense knowledge (as we show for the \textit{NE}
part of CBT) would be more limited. 

\paragraph{Knowledge Source.} As a source of common-sense knowledge we use the \textit{Open Mind Common Sense} 
part of ConceptNet 5.0 
that contains 630k fact triples. We refer to this entire source as  \textit{CN5All}. 
We conduct experiments with subparts of this data: \textit{CN5WN3} which is the WordNet 3 part of \textit{CN5All} (213k triples) 
and \textit{\CNSel}, which excludes the following WordNet relations: \emph{RelatedTo, IsA, Synonym, SimilarTo, HasContext}.

\section{Related Work}
\label{sec:related-work}

\paragraph{Cloze-Style Reading Comprehension.} Following the original 
MCTest \cite{Richardson2013-mctest-dataset} dataset 
\cam{multiple-choice}
version of cloze-style RC) 
recently several large-scale, automatically generated datasets for cloze-style reading comprehension gained a lot of attention, among others the `CNN/Daily Mail' \cite{Hermann2015-rc-cnn-dm,Onishi2016-rc-whodidwhat} and \cam{the} Children's Book Test (CBTest) data set \cite{Hill2016-booktest}.
Early work introduced simple but good \emph{single turn models} 
\cite{Hermann2015-rc-cnn-dm,Kadlec2016-as-reader,Chen2016-stanford-reader}, 
that read the document once with the question representation `in mind' and 
select an answer from 
a given set of candidates. 
More complex models 
\cite{Weston2015-memorynetworks,Dhingra2016-ga-read,Cui2016-rc-att-over-att,Munkhdalai2016-nse,Sordoni2015}
perform 
 \emph{multi-turn reading} of the story context and the question, before inferring the correct answer or use features (GA Reader, \citet{Dhingra2016-ga-read}. 
Performing multiple hops and \emph{modeling a deeper relation between question and document} was further developed by several models \cite{Seo2017-bidaf,Xiong2016-dcn-salesforce,Wang2016-rc-multi-match,Wang2016-r-net,Shen2017-reasonet} on another generation of RC datasets,
e.g.\ SQuAD \cite{Rajpurkar2016-squad}, NewsQA \cite{Trischler2017-rc-newsqa} or
TriviaQA \cite{joshi-EtAl:2017:Trivia-qa}. 

\paragraph{Integrating Background Knowledge in Neural Models.}

Integrating background knowledge in a neural model was 
proposed in the \textit{neural-checklist model} by \citet{Kiddon2016-checklist} for text generation of recipes. They 
copy words from a list of ingredients instead of 
inferring the word from a global vocabulary. 
\citet{AhnEtAl2016-Bengio-neural-knowledge-LM} proposed a \cam{language} model that 
copies fact attributes from a topic knowledge memory. 
The model predicts a fact 
in the knowledge memory using a gating mechanism
and 
given this fact, the next word to be selected is copied from the fact attributes. The knowledge facts are encoded using embeddings obtained using \textit{TransE} \cite{Bordes2013-transe}.
\citet{Dyer2017-reference-aware-language-model} extended a \textit{seq2seq} model with attention to external facts
 for dialogue and recipe generation and a co-reference resolution-aware language model. 
A similar model was adopted by \citet{He2017-seq2seq-know-2} for answer generation in 
dialogue.
Incorporating external knowledge in a neural model has 
proven
beneficial for several other tasks: \citet{Yang-and-Mithcel-LSTM-knowledge} incorporated knowledge directly into the \textit{LSTM} cell 
state 
to improve event and entity extraction.
They used knowledge embeddings trained on WordNet \cite{WordNet-Miller1990} and NELL \cite{Mitchell2015} using the \textit{BILINEAR} \cite{Yang2014-bilinear-kbc} model. 

Work similar to ours is by \citet{Long2017-world-know-rare-entity-rc}, who have introduced a new task of Rare Entity Prediction. The task is to read a paragraph from WikiLinks \cite{singh12:wiki-links} and to fill a blank field in place of a missing entity. Each missing entity is characterized with a short description derived from Freebase, and the system needs to choose one from a set of pre-selected candidates to fill the field. 
While the task is superficially similar to cloze-style reading comprehension, it differs considerably: first, when considering the text without the externally provided entity information, it is clearly ambiguous. In fact, the task is more similar to Entity Linking tasks in the Knowledge Base Population (KBP) tracks at TAC 2013-2017, which aim at detecting specific entities from Freebase. Our work, by contrast, examines the impact of injecting external knowledge in a reading comprehension, or NLU task, where the knowledge is drawn from a commonsense knowledge base, ConceptNet in our case. 
Another difference is that in their setup, the reference knowledge for the candidates is explicitly provided as a single, fixed \cam{set of} knowledge fact\cam{s} (the entity description), encoded in a single representation. In our work, we are retrieving 
\cam{(typically) distinct sets of knowledge facts}
that might (or might not) be relevant for understanding the story and answering the question. Thus, in our setup, we crucially depend on the ability of the attention mechanism to retrieve relevant  pieces of knowledge. Our aim is to examine to what extent commonsense knowledge can contribute to and improve the cloze-style RC task, that in principle is supposed to be solvable without explicitly given additional knowledge. We show that by integrating external commonsense knowledge we achieve clear improvements in reading comprehension performance over a strong baseline, and thus we can speculate that humans, when solving this RC task, are  similarly using 
commonsense knowledge as implicitly understood background knowledge. 

Recent unpublished work in \citet{Weissenborn17-knowledge} 
is driven by similar intentions. 
The authors exploit knowledge from ConceptNet to improve the performance of a reading comprehension model, experimenting on the recent SQuAD \cite{Rajpurkar2016-squad} and TriviaQA \cite{joshi-EtAl:2017:Trivia-qa} datasets. While the source of the background knowledge is the same, the way of integrating this knowledge into the model and task is different. (i) We are using attention to select unordered fact triples using key-value retrieval and (ii) we integrate the knowledge that is considered relevant explicitly for each token in the context. The model of \citet{Weissenborn17-knowledge}, by contrast, explicitly reads the acquired additional knowledge sequentially after reading the document and  question, but transfers the background knowledge implicitly, by refining the word embeddings of the words in the document and the question with the words from the supporting knowledge that share the same lemma. 
In contrast to the implicit knowledge transfer of  \citet{Weissenborn17-knowledge}, 
our explicit attention over external knowledge facts can deliver insights about the used knowledge 
and how it interacts with specific context tokens
\cam{(see Section \ref{sec:discussion})}.

\section{Experiments and Results}
\label{sec:experiments-and-results}

We perform quantitative analysis through experiments. We study the impact of the used knowledge and different model components that employ the external knowledge.
Some of the experiments below focus only on the \cam{\textit{Common Nouns (CN)}} dataset, as it has been shown to be \cam{more} challenging \cam{than \textit{Named Entities (NE)}}  \cam{in prior work}.

\subsection{Model Parameters}
We experiment with different model parameters.
\label{sec:experiments-and-results:params}

\paragraph{Number of facts.} 
\cam{We explore different sizes of knowledge memories, in terms of number of acquired facts.}
If not stated otherwise, 
we use 50 facts per example.

\paragraph{Key-Value Selection Strategy.} 
We use two strategies \cam{for defining key and value} (Key/Value): \textit{Subj/Obj} and \textit{Obj/Obj}, where \textit{Subj} and \textit{Obj} are the subject and object attributes in the fact triples and they are selected as \textit{Key}  and \textit{Value} for the KV memory \textit{(see Section \ref{sec:method:neural-model}, Querying the Knowledge Memory}). If not stated otherwise, 
we use the \textit{Subj/Obj} strategy. 

\paragraph{Answer Selection Components.} If not stated otherwise, we use ensemble attention $\alpha_{ensemble}$ (combinations of \textit{ctx} and \textit{ctx+kn}) to rank the answers. We call this our \textit{Full model} (see Sec.\ \ref{sec:method:neural-model}).

\paragraph{\cam{Hyper-parameters.}}
\cam{For our experiments we use pre-trained Glove \cite{Pennington2014-glove} embeddings, $BiGRU$ with hidden size 256, batch size of 64 and learning reate of $0.001$ as they were shown \cite{Kadlec2016-as-reader} to perform good on the AS Reader. }

\subsection{Empirical Results}
\label{sec:experiments-and-results:results}
\cam{We perform experiments with the different model parameters described above. We report accuracy on the \textit{Dev} and \textit{Test} and use the results on \textit{Dev} set for pruning the experiments.} 

\paragraph{Knowledge Sources.}
\begin{table}[]
\centering
\scalebox{0.85}{
\begin{tabular}{lll}
\bf Source  & \bf Dev   & \bf Test  \\\hline
%WordNet 3 
CN5All & 	71.40 &	66.72 \\
CN5WN3 (WN3)    & 70.70 & 68.48 \\
\CNSelected & \textbf{71.85} & 67.64\\\hline

\end{tabular}
}
\caption{Results with different knowledge sources, for CBT-CN (Full model, 50 facts).}
\label{table:results-different-sources}
\vspace{-2mm}
\end{table}
We experiment with different configuration of \textit{ConceptNet} facts (see Section \ref{sec:data}).  Results 
%from our experiments 
on the \textit{CBT CN} dataset are shown in Table \ref{table:results-different-sources}. \textit{\CNSel} 
works best on the \textit{Dev} set but \textit{CN5WN3} works much better on \emph{Test}. 
Further experiments 
use the \textit{\CNSel} setup.  

\paragraph{Number of facts.} We further experiment with different numbers of facts on the \emph{Common Nouns} dataset (Table \ref{table:results-number-of-facts}).
The best result on the \textit{Dev} set is for 50 facts \cameraready{so we use it for further experiments.}
\begin{table}[tbp]
\centering
\scalebox{0.85}{
\begin{tabular}{ccccc}
\textbf{\# facts} & \textbf{50} & \textbf{100} & \textbf{200} & \textbf{500} \\ \hline
Dev      & \textbf{71.85}       & 71.35         & 71.40         & 71.20         \\
Test    & 67.64       & 67.44        & \textbf{68.12}        & 67.24        \\ \hline
\end{tabular}
}
\caption{Results for CBT (CN)
with different numbers of facts. (Full model, \CNSel)}
\label{table:results-number-of-facts}
\vspace{-2mm}
\end{table}

\paragraph{Component ablations.}
We ensemble the attentions 
from different combinations
of the interaction between the question and document \textit{context (ctx)} representations and \textit{context+knowledge (ctx+kn)} representations in order to infer the right answer (see Section \ref{sec:method:neural-model}, Answer Ranking).  
\begin{table}[ht!]
\centering
\scalebox{0.82}{
\begin{tabular}{@{}lcccc@{}}
                        & \multicolumn{2}{c}{\textbf{NE}} & \multicolumn{2}{c}{\textbf{CN}} \\
\textbf{$ D_{repr}$ to $Q_{repr}$ interaction}    & \textbf{Dev}         & \textbf{Test}        & \textbf{Dev}        & \textbf{Test}       \\ \hline
$ D_{ctx}$, $Q_{ctx}$ (w/o know) & 75.50 & 70.30 & 68.20 & 64.80 \\ \hline
$ D_{ctx+kn}$, $Q_{ctx+kn}$ & 76.45 & 69.68 & 70.85 & 66.32 \\
 $ D_{ctx}$, $Q_{ctx+kn}$ & {\bf 77.10} & 69.72 & 70.80 & 66.32 \\
 $ D_{ctx+kn}$, $Q\cam{_{ctx}}$ & 75.65 & {\bf 70.88} & 71.20 & {\bf 67.96} \\ \hline
 Full model & 76.80 & 70.24 & {\bf 71.85} & 67.64 \\ \hline
w/o $ D_{ctx}$, $Q_{ctx}$ & 75.95 & 70.24 & 70.65 & 67.12  \\
w/o $ D_{ctx+kn}$, $Q_{ctx+kn}$ & 76.20 & 69.80 & 70.75 & 67.00 \\
w/o $ D_{ctx}$, $Q_{ctx+kn}$ & 76.55 & 70.52 & 71.75 & 66.32 \\
w/o $ D_{ctx + kn}$, $Q\cam{_{ctx}}$ & 76.05 & 70.84 & 70.80 & 66.80 \\ \hline

\end{tabular}
}
\caption{Results for different combinations of
%with different 
interactions between  document (D) and question (Q) \textit{context (ctx)} and \textit{context + knowledge (ctx+kn)} representations. (\CNSel, 50 facts)
}
\label{table:results-know-ablations}
\vspace{-2mm}
\end{table}

Table \ref{table:results-know-ablations} shows that the combination of different interactions between \textit{ctx} and \textit{ctx+kn} representations leads to \cam{clear} improvement over the \textit{w/o knowledge} setup\cam{, in particular for the \textit{Common Nouns} dataset}.
We \cam{also} performed ablations for \cam{a} model with 100 facts (see
$\Supplement$).  

\begin{table}[t]%[tbp!]
\centering
\scalebox{0.80}{
\begin{tabular}{llcccc}
           & \multicolumn{2}{c}{\textbf{NE}} & \multicolumn{2}{c}{\textbf{CN}} \\
\textbf{Key/Value} &  \textbf{Dev}         & \textbf{Test}        & \textbf{Dev}        & \textbf{Test}       \\\hline
Subj/Obj        &  76.65                & 71.52                & 71.85 & 	67.64              \\
Obj/Obj      & 76.70                & 71.28                & 71.25               & 67.48               \\\hline
\end{tabular}}
\caption{Results for key-value knowledge retrieval and 
%knowledge 
integration. (\CNSel, 50 facts). \textit{Subj/Obj} means: we attend over the fact subject (Key) and \cam{take} the weighted fact object \cam{as} value (Value).}
\label{table:results-kn-gate-kn-att}
\vspace{-4mm}
\end{table}

\paragraph{Key-Value Selection Strategy.} 
\cam{Table \ref{table:results-kn-gate-kn-att} shows that} 
for the \emph{NE} dataset, the two strategies 
\cam{perform equally well}
on the \emph{Dev} set, whereas the \emph{Subj/Obj} strategy works slightly better on the \emph{Test} set. For \emph{Common Nouns}, \emph{Subj/Obj} is 
better.

\begin{table}[bt!]
\centering
\scalebox{0.78}{
\begin{tabular}{@{}lcccc@{}}
\textbf{} & \multicolumn{2}{c}{\textbf{NE}} & \multicolumn{2}{c}{\textbf{CN}} \\
\multicolumn{1}{c}{\textbf{Models}} & \textbf{dev} & \textbf{test} & \textbf{dev} & \textbf{test} \\ \hline
Human (ctx + q) & - & 81.6 & - & 81.6 \\ \hline
\multicolumn{5}{c}{Single interaction} \\ \hline
LSTMs (ctx + q) \cite{Hill2016-booktest} & 51.2 & 41.8 & 62.6 & 56.0 \\
AS Reader & 73.8 & 68.6 & 68.8 & 63.4 \\
AS Reader (our impl) & 75.5 & 70.3 & 68.2 & 64.8  \\
% KnReader (ours) & 77.4 & 71.4 & 71.4 & 68.1 \\
KnReader (ours) & 77.4 & 71.4 & 71.8 & 67.6 \\
%rc_v15n_c4_sckn_qc_gpu07_dev3_datacbt_NE_gl.6B.100d_rnnGRU_l1_hs256_ep60_bs64_lr0.0001_treTrue_sheFalse_attTrue_lstmrFalse_cASTrue_srn1_cA2A1False_kg_avg_f100_17-08-20-12-39-16
% 
\hline
\multicolumn{5}{c}{Multiple interactions} \\ \hline
MemNNs \cite{Weston2015-memorynetworks} & 70.4 & 66.6 & 64.2 & 63.0 \\
EpiReader \cite{Trischler20160607}& 74.9 & 69.0 & 71.5 & 67.4 \\
GA Reader \cite{Dhingra2016-ga-read} & 77.2 & 71.4 & 71.6 & 68.0 \\
IAA Reader \cite{Sordoni2015} & 75.3 & 69.7 & 72.1 & 69.2 \\
AoA Reader \cite{Cui2016-rc-att-over-att} & 75.2 & 68.6 & 72.2 & 69.4 \\
GA Reader (+feat) & 77.8 & 72.0 & 74.4 & 70.7 \\
NSE \cite{Munkhdalai2016-nse}& 77.0 & 71.4 & 74.3 & 71.9 \\ 
\hline

\hline

\end{tabular}
}
\caption{Comparison of KnReader to existing end-to-end neural models on the benchmark datasets. 
}
\label{table:comparison-to-existing-work}
\vspace{-4mm}
\end{table}

\paragraph{Comparison to Previous Work.} 
Table \ref{table:comparison-to-existing-work} compares our model (\textit{Knowledgeable Reader}) to previous work on the CBT datasets. 
\cam{We show the results of our model with the settings that performed best on the \textit{Dev} sets of the two  datasets \textit{NE} and \textit{CN}: for \textit{NE},
\textit{($D_{ctx+kn}$, $Q_{ctx}$)} with 100 facts; for \textit{CN} the \textit{Full model} with \emph{50 facts}, both with \textit{\CNSel}.}

\cam{Note} that our work focuses on the impact of external knowledge and employs a \emph{single interaction (single-hop)} between the document context and the question so we \cam{primarily compare to and aim} at improving over similar models. \cam{\textit{KnReader} clearly outperforms prior single-hop models on both datasets.} 
While we do not improve over the state of the art, 
our model 
stands well among other models \cam{that perform multiple hops}. 
In the \emph{\Supplement} we also give comparison to ensemble models and some models that use 
re-ranking strategies.

\section{Discussion and Analysis}
\label{sec:discussion}

\subsection{Analysis of the empirical results.}
Our experiments examined 
%some 
key parameters of the KnReader. 
As expected, injection of background knowledge  \cam{yields only} small improvements over the baseline model for \emph{Named Entities}. 
However, on this dataset our single\cam{-hop} model is competitive to most multi-hop neural architectures. 

The integration of knowledge clearly
helps for the \emph{Common Nouns} task.
The impact of knowledge sources (Table \ref{table:results-different-sources}) is different on the \emph{Dev} and \emph{Test} sets which indicates that either the model or the data subsets are sensitive to different knowledge types and retrieved 
knowledge. 
Table \ref{table:results-kn-gate-kn-att} shows that attending over the \emph{Subj} of the knowledge triple is slightly better than \emph{Obj}. This  shows that 
using a \emph{Key-Value} memory is valuable. A reason for lower performance of \emph{Obj/Obj} is that the model picks 
\cam{facts that are similar} to
\cam{the candidate tokens,}
not adding much new information.
From the empirical results we 
see that training and evaluation with less facts is slightly better. We hypothesize that this is related to the lack of supervision on the retrieved and attended knowledge.

\subsection{Interpreting Component Importance}
\begin{figure}[!ht]
 \centering
%\hspace{-0.03\textwidth}
%10 percent%
\minipage{0.42\textwidth}
  \centering
  \includegraphics[width=\linewidth]{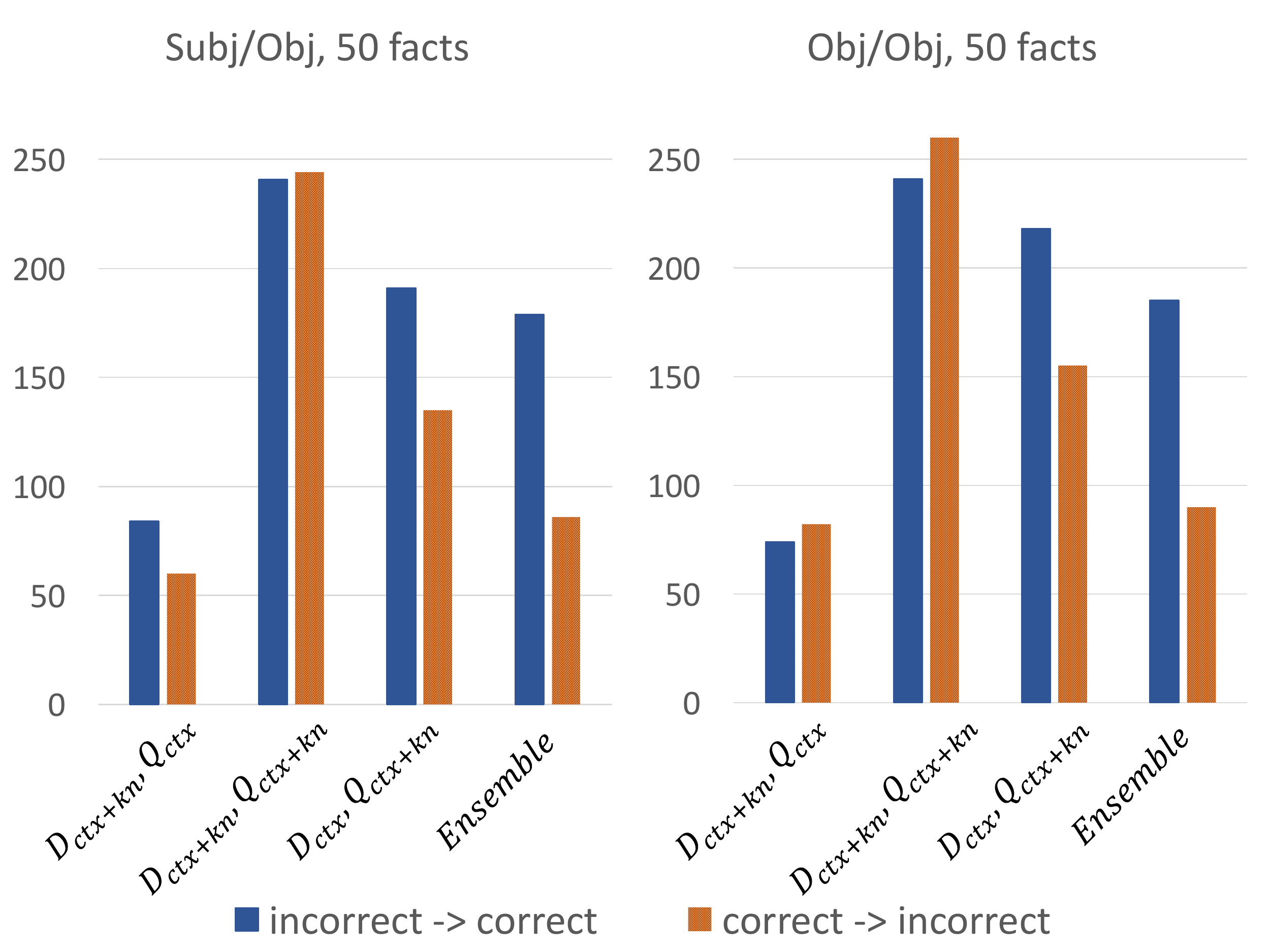}
%  \includegraphics[width=\linewidth]{figs/2018-02-22-stat-reversed-results-f50.pdf}
  %\caption{10 percent}
  %\label{figure:data-size-per-10}
\endminipage\hfill
\caption{\# of items with reversed prediction ($\pm$correct) for each combination of ({\em ctx+kn}, {\em ctx}) for Q and D. We report \cam{the number of} {\em wrong \cam{$\rightarrow$} correct} 
(blue) and {\em  correct \cam{$\rightarrow$} wrong} (orange) \cam{changes} when switching from score w/o knowledge to score w/ knowledge. The best model type
%s are \textit{D\cam{$_{cxt}$}} to \textit{Q\cam{$_{cxt+kn}$}} %+kn
%or \textcolor{red}{(best)?} 
is
\textit{Ensemble}. (\emph{Full model w/o $D_{ctx}$, $Q_{ctx}$})\textcolor{red}{.}}
\label{figure:stat-reverse-pred}
\vspace*{-3mm}
\end{figure}
Fig\cam{ure}
\ref{figure:stat-reverse-pred} \cam{shows the impact on prediction accuracy of} 
individual components of the \textit{Full model}, 
including the interaction between $D$ and 
\cam{$Q$}
with $ctx$ or $ctx+kn$ (w/o
$ctx$-only). The values for each component are obtained 
from the attention weights, without retraining the model. The difference between blue (left) and orange (right) values indicates how much the module contributes to the model. Interestingly, the ranking of the contribution ($D_{ctx},Q_{ctx+kn} > D_{ctx+kn},Q_{ctx} > D_{ctx+kn},$ $Q_{ctx+kn}$) corresponds to the component importance ablation on the $Dev$ set, lines 5-8, Table \ref{table:results-know-ablations}.

\subsection{Qualitative Data Investigation}
\cam{We will use the attention values of the interactions between $D_{ctx(+kn)}$ and $Q_{ctx(+kn)}$ and attentions to facts from each candidate token and the question placeholder to interpret 
how knowledge is employed to make a prediction for a single example. }

\textbf{Method: Interpreting Model Components.}
\cam{We manually inspect \cam{examples} from the \cam{evaluation sets} where \emph{KnReader} improves prediction (blue (left) category, Fig.\ \ref{figure:stat-reverse-pred}) or makes the prediction worse (orange (right) category, Fig.\ \ref{figure:stat-reverse-pred}). 
Figure \ref{fig:item_357} shows 
\cam{the question with placeholder, followed by answer candidates and their associated attention weights as assigned by the model \textit{w/o knowledge}. The matrix shows selected facts and their 
assigned
weights for the question and the candidate tokens. Finally, we show the attention weights determined by the knowledge-enhanced D to Q interactions.} 
The attention to the correct answer (\textit{head}) is low when the model considers the text alone (\textit{w/o knowledge}).
\cam{When} adding 
retrieved 
knowledge to the $Q$ only (row \textit{$ctx, ctx+kn$}) and to both $Q$ and $D$ (row \textit{$ctx+kn, ctx+kn$}) the score improves, while when adding knowledge to $D$ alone (row \textit{$ctx+kn, ctx$}) \cam{the score remains ambiguous}.
The combined score \textit{Ensemble} (see Eq. \ref{eq:ensemble-att}) then takes the final decision for the answer. 
In this example, the question can be answered without the story. The model tries to find 
%someting 
\cam{knowledge that is 
%close 
related
to \textit{eyes}.}
The fact \textit{eyes /r/PartOf head} \cam{is not contained} in the 
%selected 
\cam{retrieved}
knowledge but \cam{instead} 
the model selects the fact \textit{ear /r/PartOf head} which 
%is the fact with
\cam{receives} the highest attention from $Q$. The weighted \cam{\textit{Obj}} representation (\textit{head}) is added to the question with the highest weight, \cam{together with \textit{animal} and \textit{bird} from the next highly weighted facts} 
\notedone{subject head added to question: where do we see this?? Also: you don't mention the high attention to head/r/PartOf animal} 
This results in a high score for the $Q_{ctx}$ to $D_{ctx+kn}$ interaction with candidate \textit{head}. \cam{See \textit{Supplement for more details}.}

% _cbt_cloze_v15n_f50_cn5nown_tfidf_test
\begin{figure}[t]
 \centering
%\hspace{-0.03\textwidth}
%10 percent%
\minipage{0.48\textwidth}
  \centering
  \includegraphics[width=\linewidth]{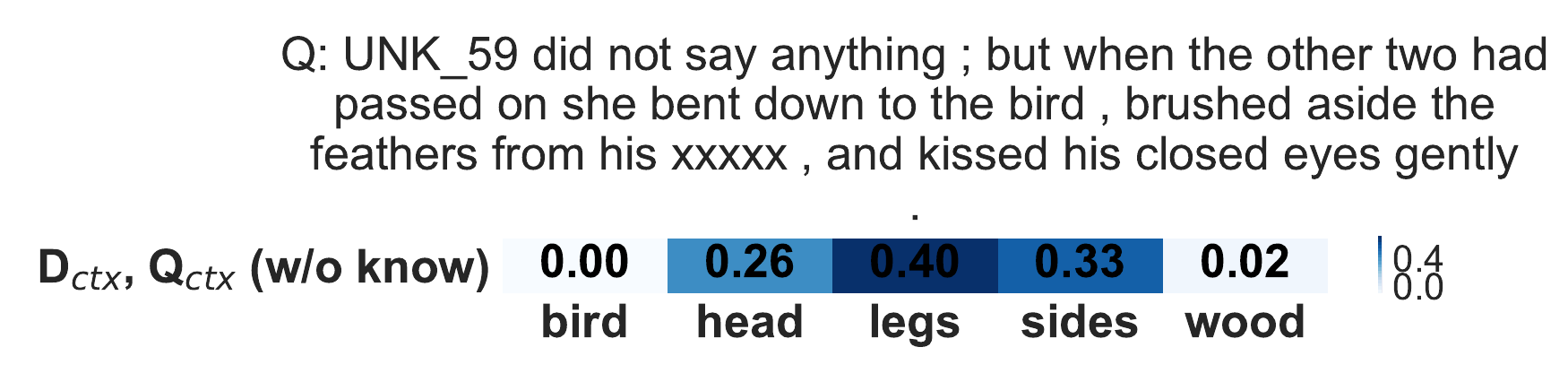}
\endminipage\hfill
\vspace*{-2mm}
\minipage{0.48\textwidth}
  \centering
  \includegraphics[width=\linewidth]{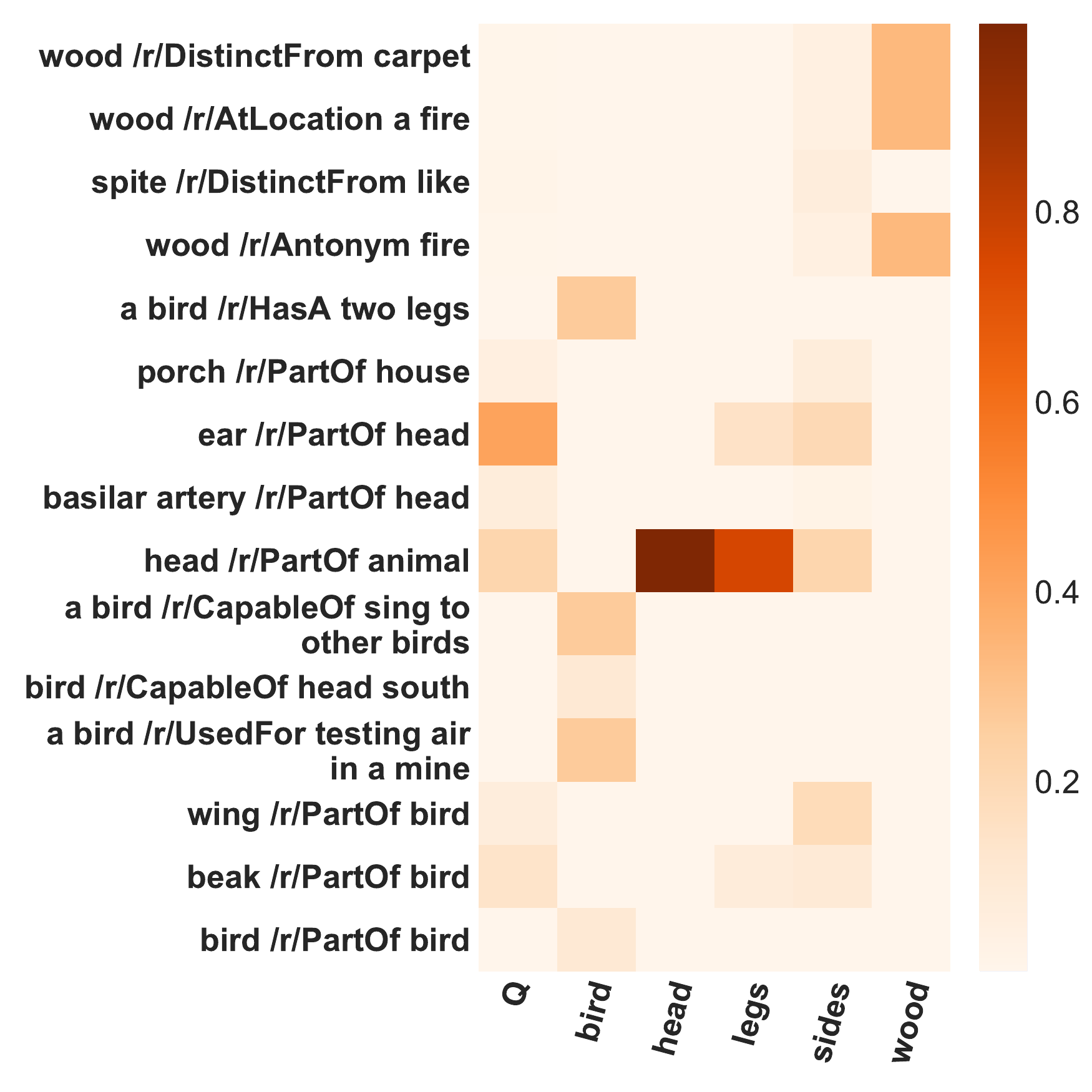}
\endminipage\hfill
\vspace*{-2mm}
\minipage{0.48\textwidth}
  \centering
  \includegraphics[width=\linewidth]{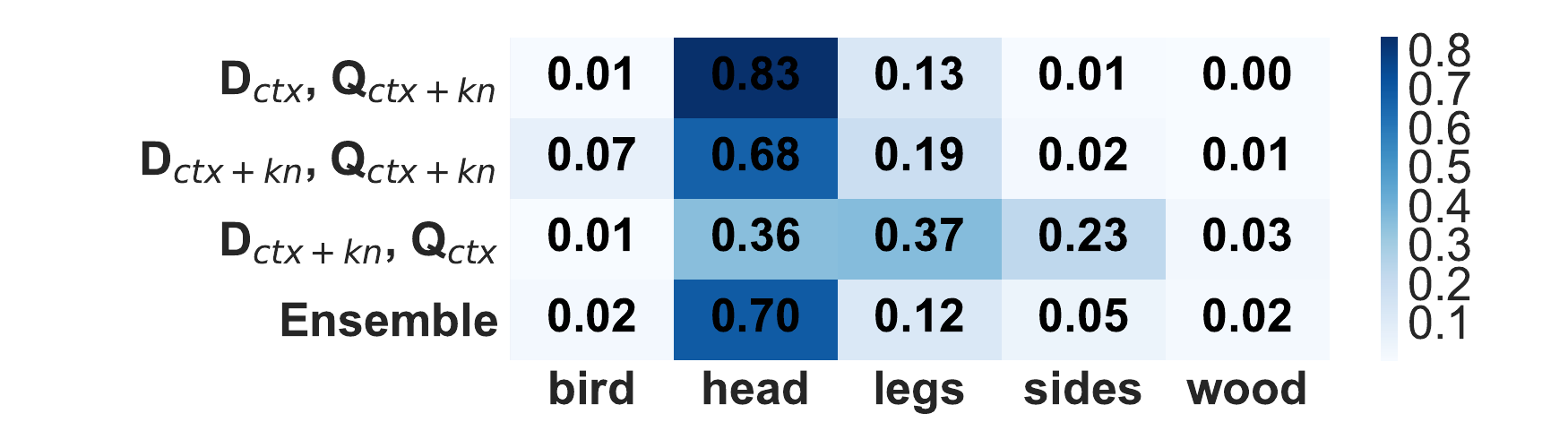}
\endminipage\hfill
\vspace*{-2mm}
\caption{Interpreting the components of \emph{KnReader}. Adding 
%retrieved 
knowledge to $Q$ and $D$ 
%helps the model to 
increases the score for the correct answer. Results for top 5 candidates are shown. (\emph{Full model, CN data, \CNSel, Subj/Obj, 50 facts})}
%, CN5NoWN
\label{fig:item_357}
\vspace*{-3mm}
\end{figure}}
\cam{Using the method described above,} 
we
analyze several example cases (presented in \emph{\Supplement}) that highlight different aspects of our model.  
Here we summarize our 
observations.

\textbf{(i.) Answer prediction from Q or Q+D.} In both human and machine RC, questions can be answered based on the question alone (Figure \ref{fig:item_357}) or jointly with the story context (Case 2, \cam{\textit{Suppl.}}). We show that empirically, enriching the question with knowledge is crucial for the first type, while enrichment of Q and D is required for the second.

\textbf{(ii.) Overcoming frequency bias.}. We show that when appropriate knowledge is available and selected, the model is able to correct a frequency bias towards an incorrect answer \camtwo{(Cases 1 and 3)}.

\textbf{(iii.) Providing appropriate knowledge.} We observe a lack of knowledge regarding events 
\camtwo{(e.g. \textit{take off} vs.\ \textit{put on clothes}, Case 2; \textit{climb up}, Case 5).}
Nevertheless relevant knowledge from CN5 \camtwo{can help}
predicting infrequent candidates \camtwo{(Case 2)}.

\textbf{(iv.) Knowledge, Q and D encoding.}
The context encoding of facts allows the model to detect knowledge that is semantically related, but not surface near to phrases in Q and D (Case \camtwo{2}).
The model finds facts to non-trivial paraphrases (e.g. \textit{undressed--naked}, Case 2).

\section{Conclusion and Future Work}
\label{sec:conclusion-and-future-work}

We \cam{propose} a neural cloze-style reading comprehension model that 
incorporates external commonsense  
knowledge, building
on a single-turn neural model. Incorporating external knowledge improves its results with a
relative 
error rate reduction of 9\% on \textit{Common Nouns}, thus the model is able to compete with more complex RC models.
We show that the types of knowledge contained in ConceptNet are useful. We provide quantitative and qualitative evidence of the effectiveness of our model, that learns how to select relevant knowledge to improve RC.
The attractiveness of our model lies in its {\em transparency and flexibility}: due to the attention mechanism, we can
trace and analyze the facts considered in answering specific questions. This opens up for deeper investigation and future improvement of RC models in a targeted way, allowing us to investigate what knowledge sources are required for different data sets and domains. 
Since our model directly integrates background knowledge with the document and question\camtwo{context} representations, it can be adapted to 
very
different task settings \cam{where we have a pair of two arguments (i.e. \textit{entailment}, 
\textit{question answering, etc.})}
In future work, we will investigate even tighter integration of the attended knowledge and
stronger reasoning methods. 

% Do not number the acknowledgment section.
%\cam{\section*{Acknowledgments}
%\vspace{-5mm}
\cam{\section*{Acknowledgments}
This work has been supported by the German Research Foundation as part of the Research Training Group Adaptive Preparation of Information from Heterogeneous Sources (AIPHES) under grant No.\ GRK 1994/1. We thank the reviewers for their helpful questions and comments.
}

\bibliography{main}
\bibliographystyle{acl_natbib}

\section*{A~ Model and \cam{Implementation} Details}
 A detailed visualization of our model, described in Section 2.2 of the main paper is shown in Fig.\ \ref{figure:knowledge-able-reader}. 
\begin{figure*}[tb!]
  \centering
  \includegraphics[width=0.98\textwidth]{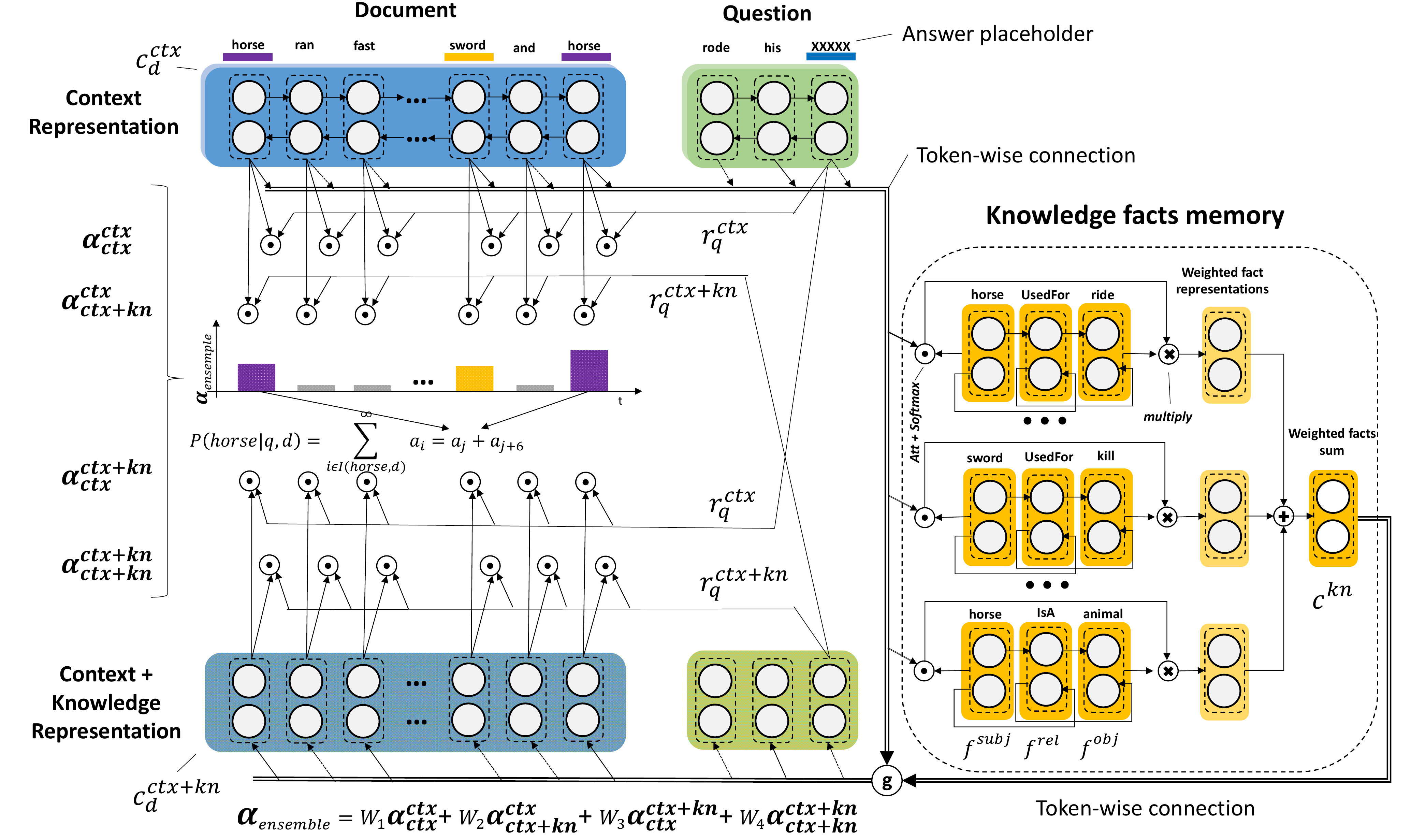}
  \caption{The Knowledgeable Reader combines plain \textit{context} \& \textit{enhanced} (\textit{context + knowledge}) repres.\ of 
  \textit{D} and \textit{Q}
 \cam{and} 
  retrieved knowledge from \cam{the} explicit memory with \cam{the} \textit{Key-Value} approach.}
  \label{figure:knowledge-able-reader}
\end{figure*}

\subsection*{Knowledge Encoding}
\begin{figure*}[t!]
  \centering
  \includegraphics[width=0.80\textwidth]{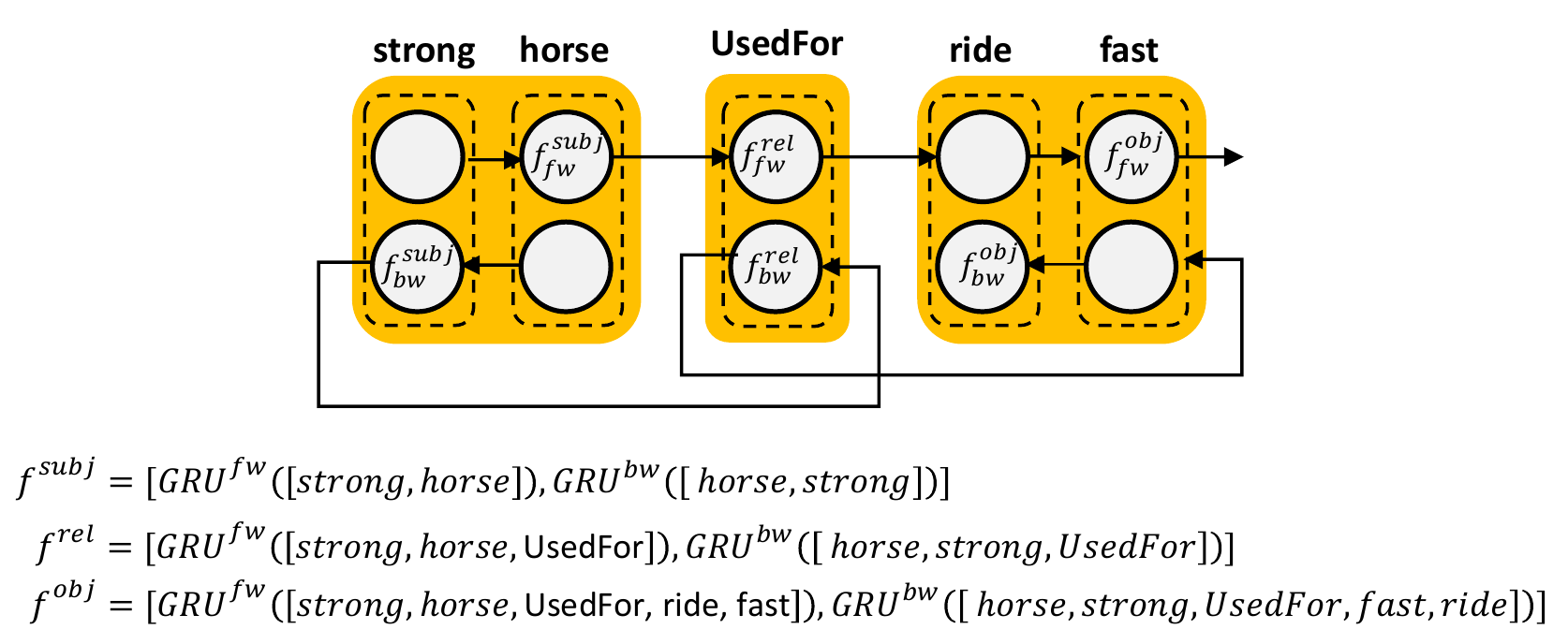}
  \caption{Encoding the knowledge triple using BiGRU. }
  \label{figure:knowledge-encoding}
\end{figure*}

We describe the fact encoding and provide comprehensive visualization of the Bi-directional GRU execution on Figure \ref{figure:knowledge-encoding}.
For each instance in the dataset, we retrieve a number of relevant facts Each retrieved fact is represented as a triple $f = (w^{subj}_{1..L_{subj}}, w^{rel}_0, w^{obj}_{1..L_{obj}})$, where $w^{subj}_{1..L_{subj}}$ and $w^{obj}_{1..L_{obj}}$ are a multi-word expressions representing the $subject$ and $object$ with 
 sequence lengths $L_{subj}$ and $L_{obj}$, and $w^{rel}_0$ is a word token corresponding to a relation.\footnote{The $0$ in $w^{rel}_0$ indicates that we encode the relation as a single \textit{relation type} word. Ex. \textit{/r/IsUsedFor}.}
 As a result of fact encoding, we obtain a separate knowledge memory for each instance in the data. 
To encode the knowledge we use 
a $BiGRU$ to encode the 
\cam{triple argument}
\cam{tokens}
into the following 
context\cam{-encoded} representations:
\begin{align}
f^{subj}_{last} = BiGRU(Emb(w^{subj}_{1..L_{subj}}), 0) \\
f^{rel}_{last} = BiGRU(Emb(w^{rel}_{0}), f^{subj}_{last})\\
f^{obj}_{last} = BiGRU(Emb(w^{obj}_{1..L_{subj}}), f^{rel}_{last})
\end{align}

, where $f^{subj}_{last}$, $f^{rel}_{last}$, $f^{obj}_{last}$ are the final hidden 
\cam{states}
of the context encoder $BiGRU$, that are also used as initial representations for the encoding of the next triple attribute in left-to-right order. 
The motivation behind this encoding is: 
(i) 
\cam{We e}ncode the knowledge fact attributes in the same vector space as the plain tokens; (ii) 
\cam{we p}reserve the triple directionality; (iii)
we use the relation type as a way of filtering the \textit{subject} information to initialize the \textit{object}. 

\subsection*{Model Implementation Parameters}
\label{sec:experiments-and-results:params}
 We implement our model in \textit{TensorFlow 0.12}  \cite{tensorflow2015-whitepaper}. Below we report pre-processing steps and hyper-parameters required for reproducing the model.

\paragraph{Dataset.} We perform experiments on the \textit{Common Nouns} and \textit{Named Entities} parts of the Children's Book Test (CBT) \citep{Hill2016-booktest}. \footnote{The dataset can be downloaded from: \url{http://www.thespermwhale.com/jaseweston/babi/CBTest.tgz} } 

\paragraph{Pre-processing.} For each instance of the dataset (21 sentences, 20 for the story and 1 for  question), we remove the line number, which is originally presented in the text as a first token of the sentence and split the tokens using \textit{str.split() in Python 2.7}. We then concatenate the tokens for the sentences in the story into a single list of story tokens $d_{1..m}$. 

\paragraph{Knowledge Source.} We use knowledge from the Open Mind Common Sense (OMCS, \citet{Singh2002-common-sense-kw-omcs}) part of ConceptNet \cite{Speer2017Conceptnet55}, a crowd-sourced resource of
commonsense \cam{knowledge} 
with a total of $\sim$630k facts. 
\footnote{ConceptNet 5 github page: \url{https://github.com/commonsense/conceptnet5}.}
The exact knowledge splits required for our experiments will be available in \textit{json} format.
\footnote{Knowledge splits: \url{https://github.com/tbmihailov/enhancing-rc-with-commonsense}.}
\paragraph{Vocabulary.} To build the vocabulary we select the words that occur at least 5 times in the training set. 
We extend the vocabulary with all words retrieved from the knowledge source. 
All words are lowercased. Following \citet{Kadlec2016-as-reader} we use multiple unknown tokens (UNK$_1$, UNK$_2$, \dots, UNK$_{100}$). In each example, for each unknown word, we pick randomly an unknown token from the list and use it for all occurrences of the word in the \cam{document (story) and question}.

\paragraph{Word Embeddings.} We use Glove 100D\footnote{The embeddings can be downloaded from: \url{http://nlp.stanford.edu/data/glove.6B.zip} } word embeddings pre-trained on 6B tokens from Wikipedia and Gigaword5. We initialize the out-of-vocabulary words by sampling from a uniform distribution in range [$-$0.1, 0.1]. We optimize all word embeddings in the first 8000 training steps.

\paragraph{Encoder Hidden Size.} We use a hidden size of 256 for the \textit{GRU} encoder states \cam{(512 output for our bi-directional encoding)}. This setting has been shown to perform well for the Attention Sum Reader \cite{Kadlec2016-as-reader}. 

\paragraph{Batching, Learning rate, Sampling.} We sort the data examples in the training set by document length and create batches with 
%size of 
64 examples. For each training step we pick batches randomly. 
After every 1000 training steps we evaluate the models 
on the validation \textit{Dev} set. We train for 60 epochs and pick the model with the highest validation accuracy to make the predictions for \textit{Test}.

\paragraph{Optimization.} We use cross entropy loss on the predicted scores for each answer candidate. We use Adam \cite{Kingma2015-adam} optimizer with initial learning rate of \textit{0.001} and clip the gradients in the range [$-$10, 10].

\section*{B~ Quantitative Analysis}
\subsection*{Additional Ablation Experiments}
Due to space limitation in the main paper, we present additional results here.
In addition to ablation of model components for 50 facts, we perform experiments for 100 as well. 
\begin{table}[b!]
\centering
\scalebox{0.75}{
\begin{tabular}{@{}lcccc@{}}
                        & \multicolumn{2}{c}{\textbf{NE}} & \multicolumn{2}{c}{\textbf{CN}} \\
\textbf{$ D_{repr}$ to $Q_{repr}$ interaction}    & \textbf{Dev}         & \textbf{Test}        & \textbf{Dev}        & \textbf{Test}       \\ \hline
$ D_{ctx}$, $Q_{ctx}$ (w/o know)            & 75.50              & 70.30              & 68.20             & 64.80            \\ \hline
$ D_{ctx+kn}$, $Q_{ctx+kn}$   & 75.50           & 70.28          & 69.80           & 65.60           \\
$ D_{ctx}$, $Q_{ctx+kn}$       & 74.20            & 69.88          & 70.40           & 66.56         \\
$ D_{ctx+kn}$, $Q_{ctx}$         & \textbf{77.40  }          & 71.40            & 70.95         & 67.52         \\ \hline
All                     & 76.65          & \textbf{71.52  }        & 70.80           & 67.08         \\ \hline
w/o $ D_{ctx}$, $Q_{ctx}$        & 76.70            & 70.68          & \textbf{71.10  }         & \textbf{67.68  }       \\
w/o $ D_{ctx+kn}$, $Q_{ctx+kn}$ & 76.35          & 70.88          & 70.95         & 67.44         \\
w/o $ D_{ctx}$, $Q_{ctx+kn}$     & 76.90            & 71.32          & 70.70           & 67.12         \\
w/o $ D_{ctx + kn}$, $Q_{ctx}$       & 76.50              & 70.64          & 70.75         & 66.88         \\ \hline
\end{tabular}
}
\caption{Results for different combinations of
%with different 
interactions between  document (D) and question (Q) \textit{context (ctx)} and \textit{context + knowledge (ctx+kn)} representations. (Subj/Obj, 100 facts)}
%We report accuracy and differences to the model w/o knowledge.}
\label{table:results-know-ablations-f100}
\vspace{-2mm}
\end{table}

The results are shown in Table \ref{table:results-know-ablations-f100}. \cam{The results show a similar tendency, but in this setting, omitting the model without knowledge enrichment yields best results for the CN data.}

\subsection*{Results for Ensemble Models}
For each dataset we combine our best 11 runs \cam{and use} majority voting to predict the answer for our \emph{Ensemble} model. 

\begin{table}[h!]
\centering
\scalebox{0.78}{
\begin{tabular}{@{}lcccc@{}}
\textbf{} & \multicolumn{2}{c}{\textbf{NE}} & \multicolumn{2}{c}{\textbf{CN}} \\
\multicolumn{1}{c}{\textbf{Models}} & \textbf{dev} & \textbf{test} & \textbf{dev} & \textbf{test} \\ \hline
%Human (query) \cite{Hill2016-booktest} & - & 0.5 & - & 64.4 \\
Human (ctx + q) & - & 81.6 & - & 81.6 \\ \hline
% \multicolumn{5}{c}{Single interaction} \\ \hline
% LSTMs (ctx + q) \cite{Hill2016-booktest} & 51.2 & 41.8 & 62.6 & 56.0 \\
% AS Reader & 73.8 & 68.6 & 68.8 & 63.4 \\
% AS Reader (our impl) & 75.5 & 70.3 & 68.2 & 64.8  \\
% % KnReader (ours) & 77.4 & 71.4 & 71.4 & 68.1 \\
% KnReader (ours) & 77.4 & 71.4 & 71.8 & 67.6 \\
% %rc_v15n_c4_sckn_qc_gpu07_dev3_datacbt_NE_gl.6B.100d_rnnGRU_l1_hs256_ep60_bs64_lr0.0001_treTrue_sheFalse_attTrue_lstmrFalse_cASTrue_srn1_cA2A1False_kg_avg_f100_17-08-20-12-39-16
% % 
% \hline
% \multicolumn{5}{c}{Multiple interactions} \\ \hline
% MemNNs \cite{Weston2015-memorynetworks} & 70.4 & 66.6 & 64.2 & 63.0 \\
% EpiReader \cite{Trischler20160607}& 74.9 & 69.0 & 71.5 & 67.4 \\
% GA Reader \cite{Dhingra2016-ga-read} & 77.2 & 71.4 & 71.6 & 68.0 \\
% IAA Reader \cite{Sordoni2015} & 75.3 & 69.7 & 72.1 & 69.2 \\
% %77.2 71.4 71.6 68.0
% AoA Reader \cite{Cui2016-rc-att-over-att} & 75.2 & 68.6 & 72.2 & 69.4 \\
% GA Reader (+feat) & 77.8 & 72.0 & 74.4 & 70.7 \\
% NSE \cite{Munkhdalai2016-nse}& 77.0 & 71.4 & 74.3 & 71.9 \\ 
% \hline

\multicolumn{5}{c}{Ensemble} \\ \hline
%MemNN \cite{Hill2016-booktest} & 70.4 & 66.6 & 64.2 & 63.0 \\
AS Reader \cite{Kadlec2016-as-reader}& 74.5 & 70.6 & 71.1 & 68.9 \\
KnReader (ours)  & 78.0 & 73.3 & 72.2 & 70.6 \\
EpiReader \cite{Trischler20160607}& 76.6 & 71.8 & 73.6 & 70.6 \\
IAA Reader  \cite{Sordoni2015} & 76.9 & 72.0 & 74.1 & 71.0 \\ 
AoA Reader \cite{Cui2016-rc-att-over-att} & 78.9 & 74.5 & 74.7 & 70.8 \\
\hline
\multicolumn{5}{c}{Re-ranking} \\ \hline
%MemNN \cite{Hill2016-booktest} & 70.4 & 66.6 & 64.2 & 63.0 \\
AoA Reader (re-ranking)& 79.6 & 74.0 & 75.7 & 73.1\\
AoA Reader (ens + re-rank)& 80.3 & 75.6 & 77.0 & 74.1 \\
\hline

\end{tabular}
}
\caption{Comparison of \textit{KnReader}
%Knowledgeable Reader 
to existing ensemble models and models that use re-ranking. 
%\anettenote{if we need space, maybe set references in small/fn/tiny size and use less wide columns for numbers}
}
\label{table:comparison-to-existing-work-ensemble}
\end{table}

In Table \ref{table:comparison-to-existing-work-ensemble} we show the comparison with multi-hop models. We report \textit{Accuracy} on the \textit{Dev} and \textit{Test} sets, rounded to the first decimal point as done in previous work.
The \emph{AoA Reader} \cite{Cui2016-rc-att-over-att} uses re-ranking as a post-processing step \cam{and} the other neural models are not directly comparable.

\section*{C~ Manual Analysis and Visualization}
\paragraph{Case 1} We
provide an extended illustration of the example discussed in the main paper in Figure  \ref{fig:item_357}. 
% _cbt_cloze_v15n_f50_cn5nown_tfidf_test
\begin{figure*}[!t]
 \centering
%\hspace{-0.03\textwidth}
%10 percent%
\minipage{0.95\textwidth}
  \centering
   % ITEM 357
   \input{supl_analysis_figures/item_357_story.tex}
\endminipage\hfill
\vspace*{2mm}
\minipage{0.95\textwidth}
  \centering

  \includegraphics[width=\linewidth]{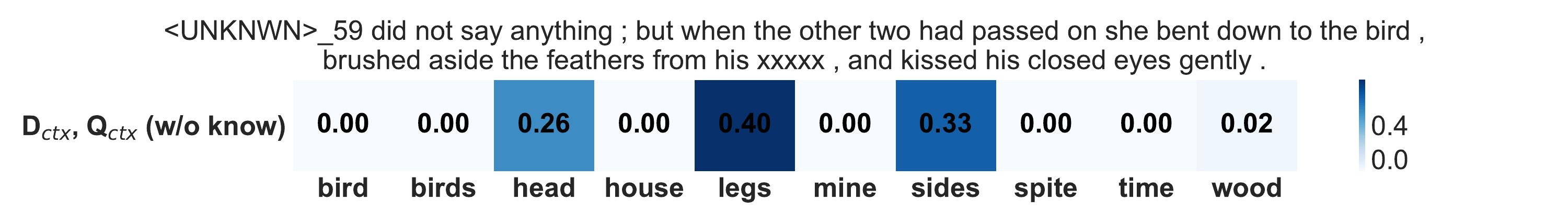}
\endminipage\hfill
\vspace*{-2mm}
\minipage{0.95\textwidth}
  \centering
  \includegraphics[width=\linewidth]{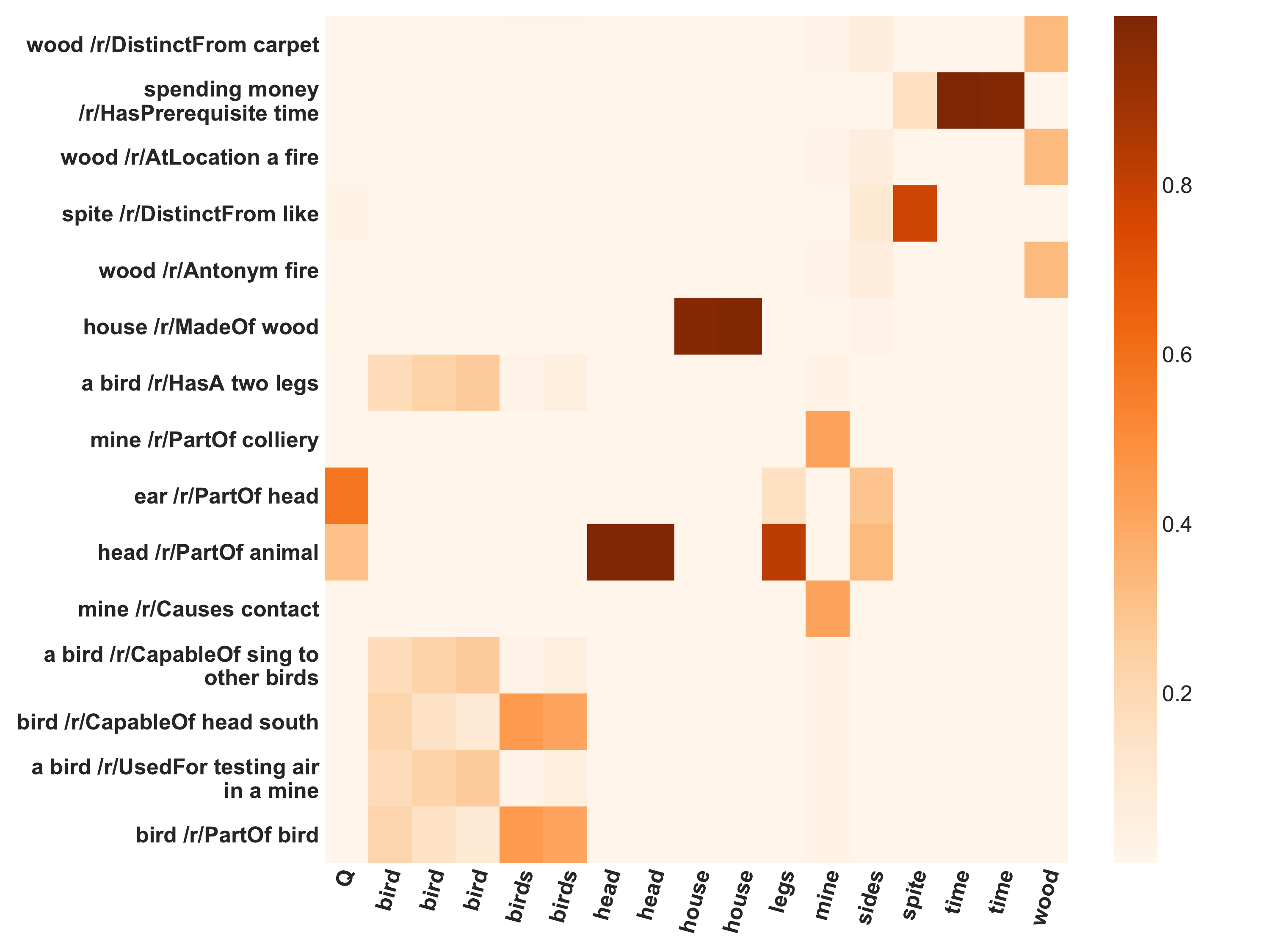}
\endminipage\hfill
\vspace*{-2mm}
\minipage{0.95\textwidth}
  \centering
  \includegraphics[width=\linewidth]{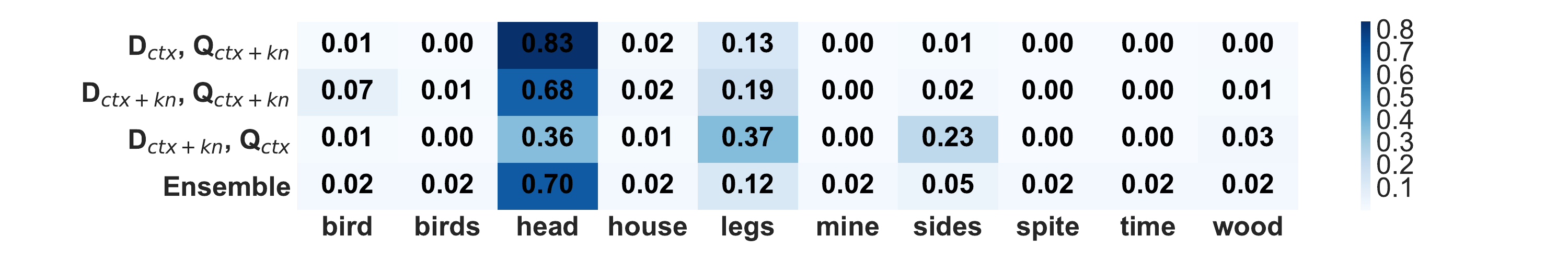}
\endminipage\hfill
\caption{\textbf{Case 1:} Interpreting the components of \emph{KnReader} (\emph{Full model}). Adding retrieved knowledge to $Q$ and $D$ helps the model to increase the score for the correct answer. Results for top 5 candidates are shown.
(Subj/Obj as key-value memory, 50 facts, \CNSel) (Item \#357)}
\label{fig:item_357}
%\vspace*{-3mm}
\end{figure*}

We manually inspect \cam{examples} from the \cam{evaluation sets} where \emph{KnReader} improves prediction or makes the prediction worse . %\todo{Insert what is now in fn 5 into the caption of Fig. 4} 
%\footnote{We are using the \textit{Full model} trained on the 
%\textit{Common Nouns} dataset with \textit{CN5NoWN} knowledge. }
Figure \ref{fig:item_357} shows 
%such? values-$>$ 
\cam{the question with placeholder, followed by answer candidates and their associated attention weights as assigned by the model \textit{w/o knowledge}. The matrix shows selected facts and their learned weights for question and the candidate tokens. Finally, we show the attention weights determined by the knowledge-enhanced D to Q interactions.} 
% \todo{I have reedited the expanatory text in the main paper. If you are ok with it (and possibly revised what I ask), copy it here, too.}
% The example shows that the attention to the correct answer (\textit{head}) is low when the model considers the text alone (\textit{w/o knowledge}).
% It then retrieves knowledge and computes the attention for the updated representations.
% In the given example adding retrieved knowledge to the $Q$ only (row \textit{$ctx, ctx+kn$}) and both $Q$ and $D$ (row \textit{$ctx+kn, ctx+kn$}) improves the score while adding the knowledge to $D$ alone (row \textit{$ctx+kn, ctx$}) keeps the score still ambiguous. The combined score \textit{ensemble} 
% %(see Eq. \ref{eq:ensemble-att}) 
% then takes the final decision for the answer. 
% In this example, the question can be answered without the story. The model tries to find someting that is close to \textit{eyes} by meaning. We do not have the fact \textit{eyes /r/PartOf head} in the selected knowledge but we have the fact \textit{ear /r/PartOf head} which is the fact with the highest attention from $Q$. The subject representation (\textit{head}) is added to the question with the highest weight. This results in high score for the $Q_{ctx}$ to $D_{ctx+kn}$ interaction with candidate \textit{head}.

% from main paper
The attention to the correct answer (\textit{head}) is low when the model considers the text alone (\textit{w/o knowledge}).
\cam{When} adding retrieved knowledge to the $Q$ only (row \textit{$ctx, ctx+kn$}) and to both $Q$ and $D$ (row \textit{$ctx+kn, ctx+kn$}) the score improves, while when adding knowledge to $D$ alone (row \textit{$ctx+kn, ctx$}) \cam{the score remains ambiguous}.
The combined score \textit{Ensemble} then takes the final decision for the answer. 
In this example, the question can be answered without the story. The model tries to find 
%someting 
\cam{knowledge that is 
%close 
related
to \textit{eyes}.}
The fact \textit{eyes /r/PartOf head} \cam{is not contained} in the 
%selected 
\cam{retrieved}
knowledge but \cam{instead} 
the model selects the fact \textit{ear /r/PartOf head} which 
%is the fact with
\cam{receives} the highest attention from $Q$. The weighted \camtwo{\textit{Obj}} representation (\textit{head}) is added to the question with the highest weight, \cam{together with \textit{animal} and \textit{bird} from the next highly weighted facts} 
\notedone{subject head added to question: where do we see this?? Also: you don't mention the high attention to head/r/PartOf animal} 
This results in a high score for the $Q_{ctx}$ to $D_{ctx+kn}$ interaction with candidate \textit{head}. %\cam{See \textit{Supplement for more details}.}

\begin{figure*}[!t]
 \centering
%\hspace{-0.03\textwidth}
%10 percent%
\minipage{0.95\textwidth}
  \centering
   % ITEM 52
   \input{supl_analysis_figures/item_52_story.tex}
\endminipage\hfill
\vspace*{2mm}
\minipage{0.95\textwidth}
  \centering

  \includegraphics[width=\linewidth]{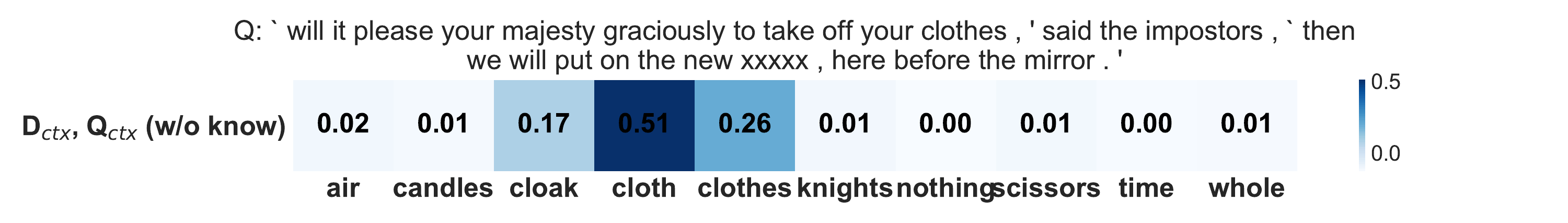}
\endminipage\hfill
\vspace*{-2mm}
\minipage{0.95\textwidth}
  \centering
  \includegraphics[width=\linewidth]{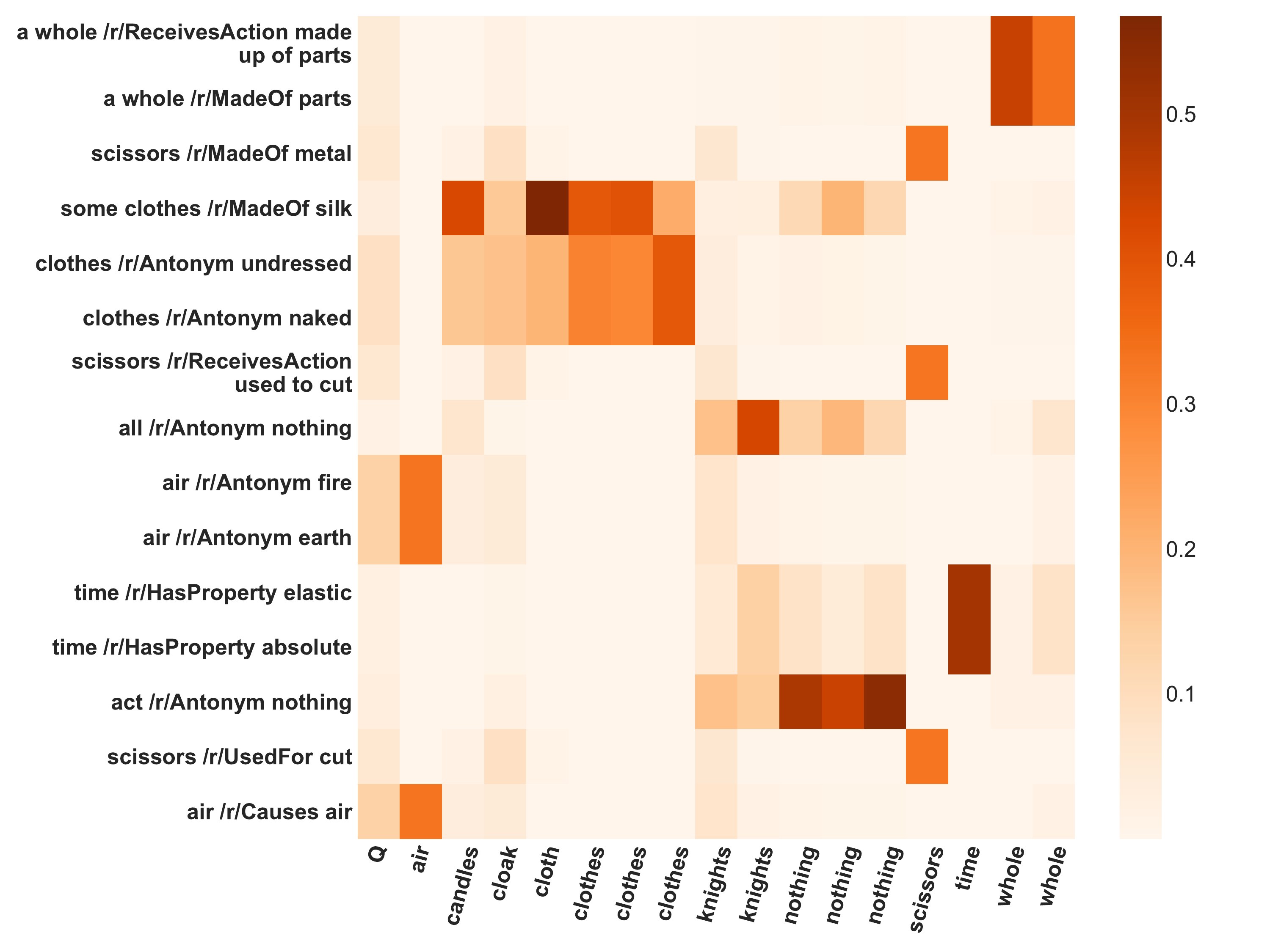}
\endminipage\hfill
\vspace*{-2mm}
\minipage{0.95\textwidth}
  \centering
  \includegraphics[width=\linewidth]{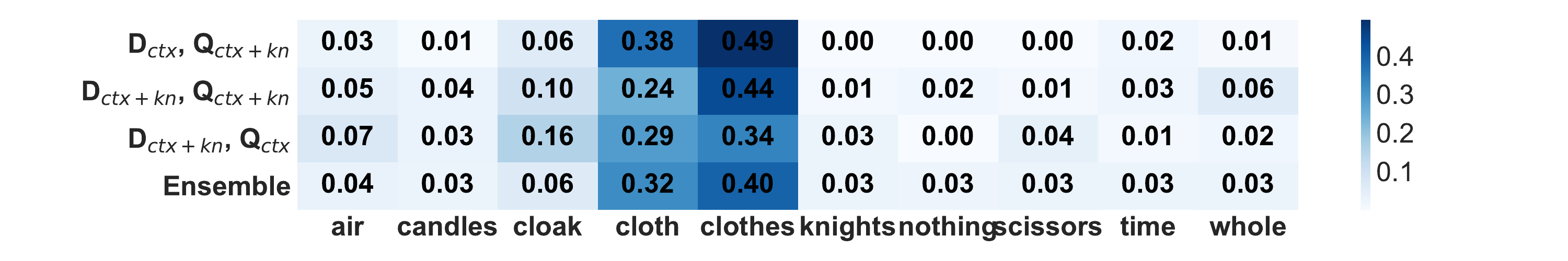}
\endminipage\hfill
\caption{\textbf{Case 2:} Interpreting the components of \emph{KnReader} (\emph{Full model}). Adding retrieved knowledge to $Q$ and $D$ helps the model to increase the score for the correct answer. Results for top 5 candidates are shown.
(Subj/Obj as key-value memory, 50 facts, \CNSel) (Item \#52)}
\label{fig:item_52}
%\vspace*{-3mm}
\end{figure*}

\paragraph{Case 2}
Figure \ref{fig:item_52} shows another interesting example. The document is part of the \textit{The kings new clothes} by Hans Christian Andersen. While, given the story, many of the choices are plausible (\textit{cloth, clothes, nothing, air, cloak}) the model without knowledge selects \textit{cloth} as the most probable answer. Adding the knowledge facts reverts the answer. We can speculate that the reason is the fact \textit{clothes /r/Antonym undressed} retrieved by the \cam{answer} candidate token \textit{clothes} which has multiple occurrences in the text,
and
\cam{since}
the updated representation combines well with the phrase \textit{put on} which is antonym to undressed \cam{\textit{clothes /r/Antonym undressed} and \textit{clothes /r/Antonym naked}}. A reason for this \cam{could} 
also be the high frequency of clothes in the story. However, the example \cam{cannot be answered using the story context} 
alone, \cam{as it} 
talks about the imaginary, not existing (\textit{air, nothing}) new clothes of the king. 

% _cbt_cloze_v15n_f50_cn5nown_tfidf_test

\cam{The example also shows what kind of knowledge is missing in our currently used resources: ideally, the question can be answered using information from the question alone, by analyzing the meaning of the phrases \textit{take off your clothes} and \textit{then we will put on the new XXXX}. If they were available, the model could exploit the knowledge that \textit{taking off (clothes)} and \textit{putting on (clothes)} are actions often performed in temporal sequence.}

\begin{figure*}[!ht]
 \centering
%\hspace{-0.03\textwidth}
%10 percent%
\minipage{0.95\textwidth}
  \centering
   % ITEM 240
   \input{supl_analysis_figures/item_240_story.tex}
\endminipage\hfill
\vspace*{2mm}
\minipage{0.95\textwidth}
  \centering

  \includegraphics[width=\linewidth]{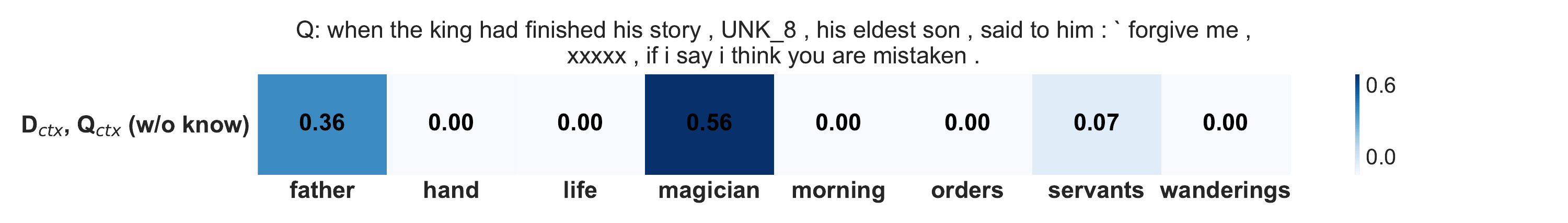}
\endminipage\hfill
\vspace*{-2mm}
\minipage{0.95\textwidth}
  \centering
  \includegraphics[width=\linewidth]{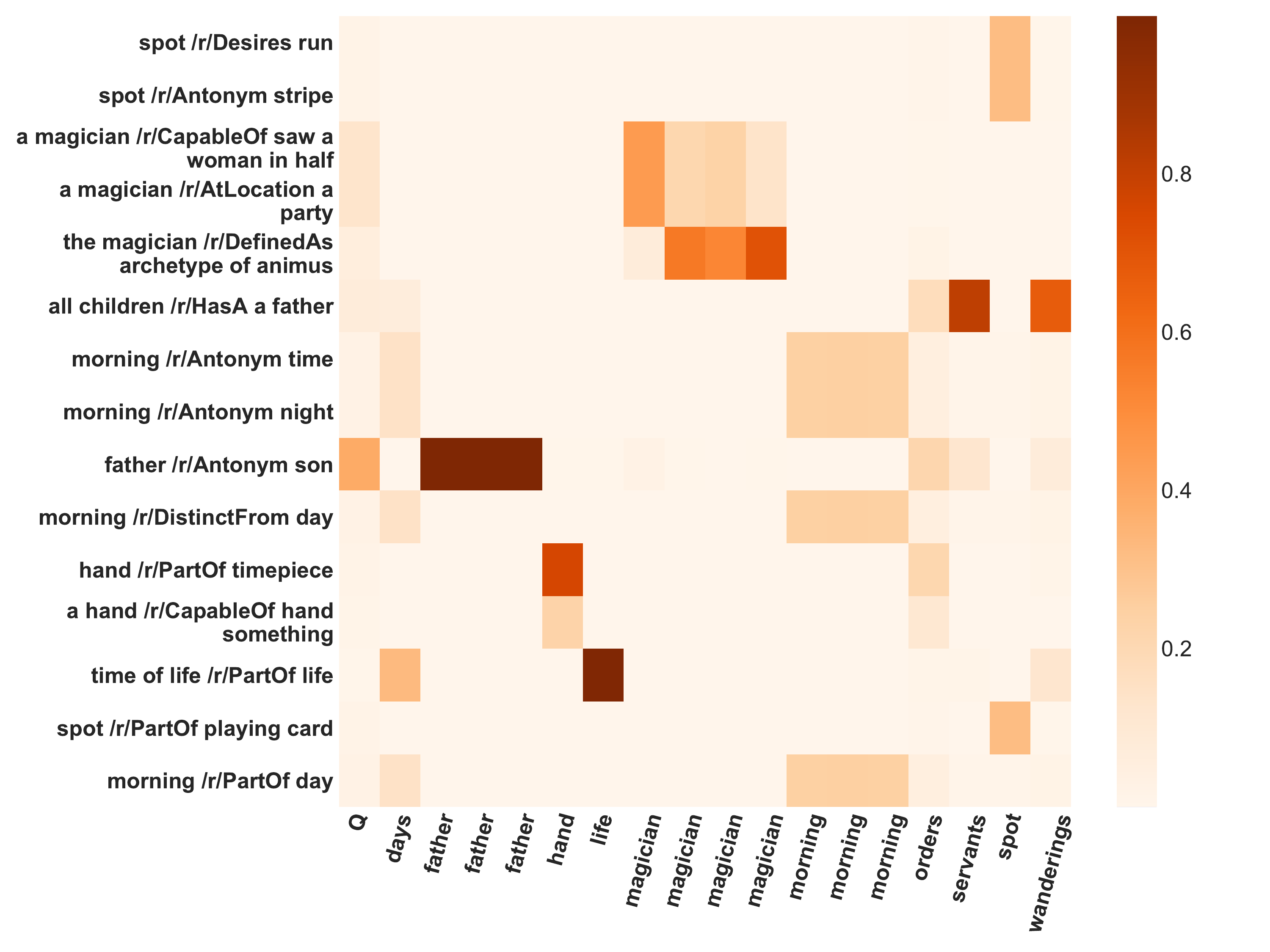}
\endminipage\hfill
\vspace*{-2mm}
\minipage{0.95\textwidth}
  \centering
  \includegraphics[width=\linewidth]{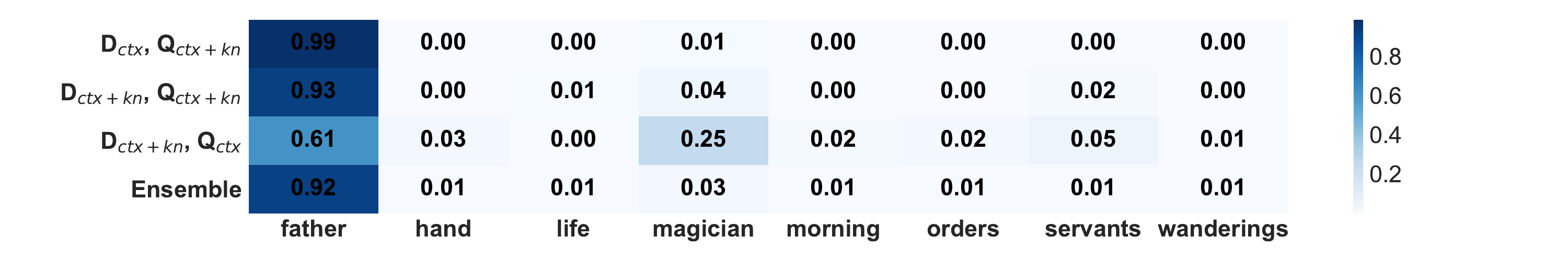}
\endminipage\hfill
\caption{\textbf{Case 3}: Interpreting the components of \emph{KnReader} (\emph{Full model}). Adding retrieved knowledge to $Q$ and $D$ helps the model to increase the score for the correct answer. Results for top 5 candidates are shown.
(Subj/Obj as key-value memory, 50 facts, \CNSel) (Item \#240)}
\label{fig:item_240}
%\vspace*{-3mm}
\end{figure*}

\paragraph{Case 3}
In Figure \ref{fig:item_240} we have an example where the model overcomes the frequency bias of the story (\textit{magician} occurs 4 times) to select a more plausible example (\textit{father}) \cam{using the fact \textit{father /r/Antonym son}}. 
% _cbt_cloze_v15n_f50_cn5nown_tfidf_test

\begin{figure*}[!ht]
 \centering
%\hspace{-0.03\textwidth}
%10 percent%
\minipage{0.95\textwidth}
  \centering
   % ITEM 172
   \input{supl_analysis_figures/item_172_story.tex}
\endminipage\hfill
\vspace*{2mm}
\minipage{0.95\textwidth}
  \centering

  \includegraphics[width=\linewidth]{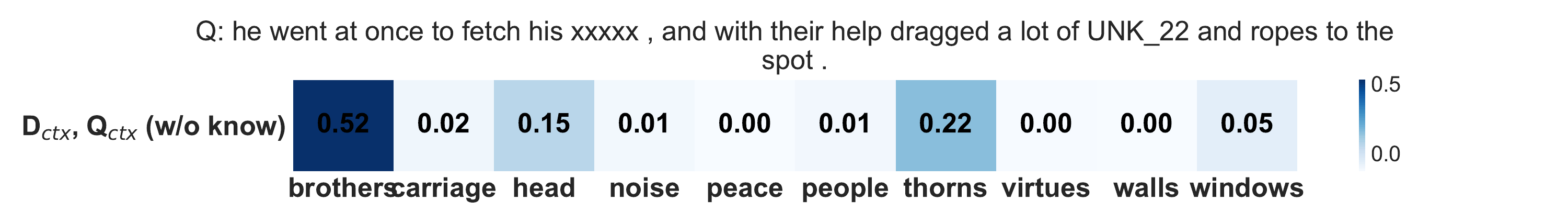}
\endminipage\hfill
\vspace*{-2mm}
\minipage{0.95\textwidth}
  \centering
  \includegraphics[width=\linewidth]{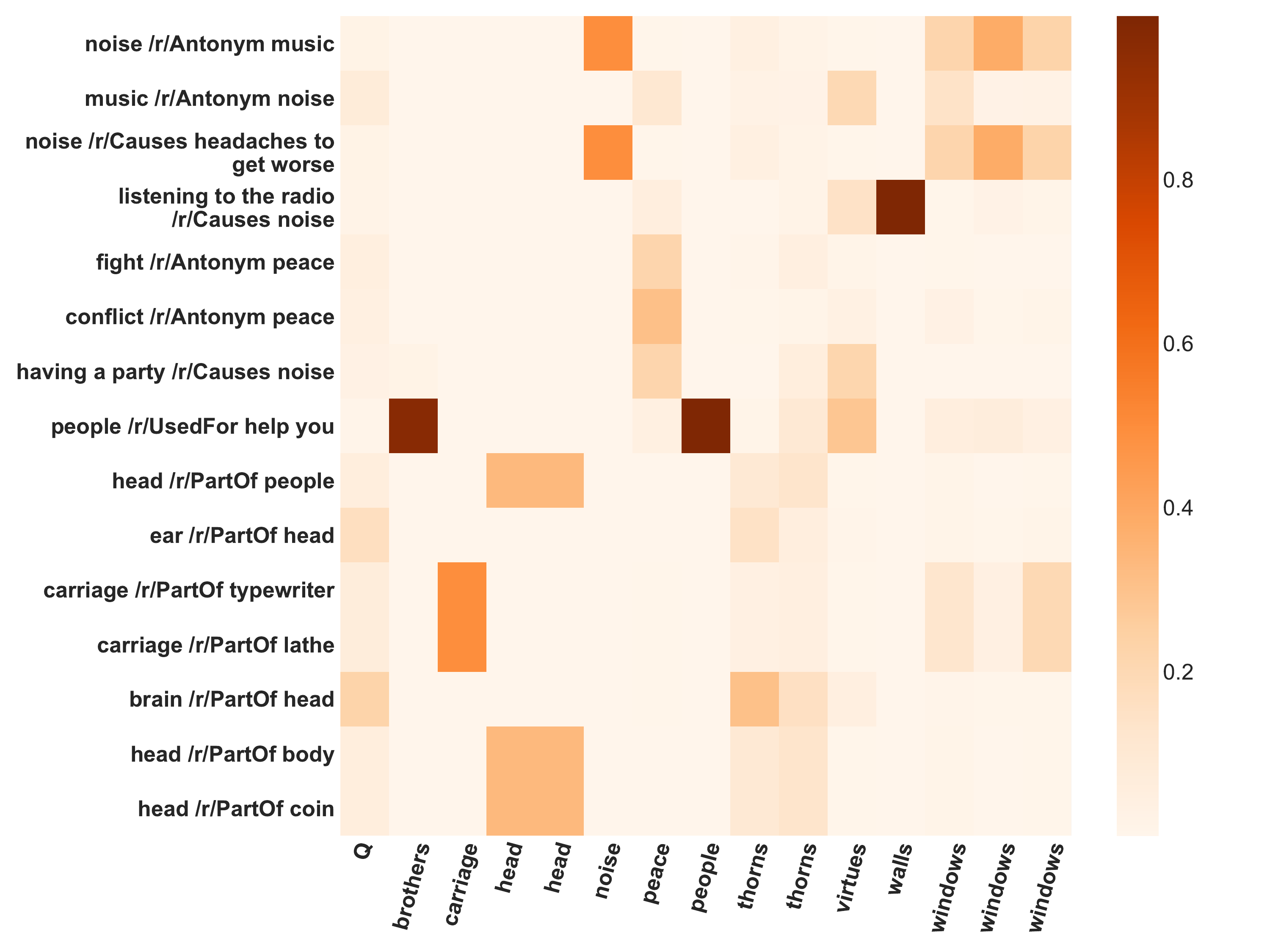}
\endminipage\hfill
\vspace*{-2mm}
\minipage{0.95\textwidth}
  \centering
  \includegraphics[width=\linewidth]{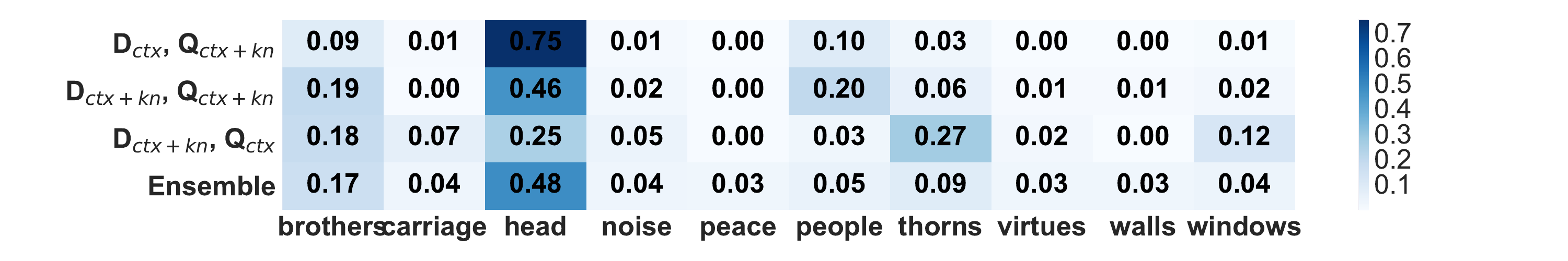}
\endminipage\hfill
\caption{\textbf{Case 4:} Interpreting the components of \emph{KnReader} (\emph{Full model}). Adding retrieved knowledge to $Q$ and $D$ confuses the model and decreases
%to increase 
the score for the correct answer. Results for top 5 candidates are shown.
(Subj/Obj as key-value memory, 50 facts, \CNSel) (Item \#172)}
\label{fig:item_172}
%\vspace*{-3mm}
\end{figure*}

\paragraph{Case 4}
Figure \ref{fig:item_172} shows an example where 
a correct initial prediction 
\cam{obtained without knowledge is reversed and a clearly wrong answer is selected instead.} 
\cam{Although a relevant fact is selected (\textit{people /r/UsedFor help you}), apparently, the model misses the information that \textit{brothers are people} and can't combine the acquired concept \textit{help you} with the question context \textit{and with their help dragged ...}}, and thus, the correct answer is not sufficiently promoted.
\notedone{not sure what you mean by what can be done with 'head' ; taken out: The model clearly misses information about what can be \textcolor{red}{concluded from} 
(\textit{fetch} your own \textit{head}) with a given concept (\textit{head}).
}
% _cbt_cloze_v15n_f50_cn5nown_tfidf_test

\begin{figure*}[!ht]
 \centering
%\hspace{-0.03\textwidth}
%10 percent%
\minipage{0.95\textwidth}
  \centering
   % ITEM 187
   \input{supl_analysis_figures/item_187_story.tex}
\endminipage\hfill
\vspace*{2mm}
\minipage{0.95\textwidth}
  \centering

  \includegraphics[width=\linewidth]{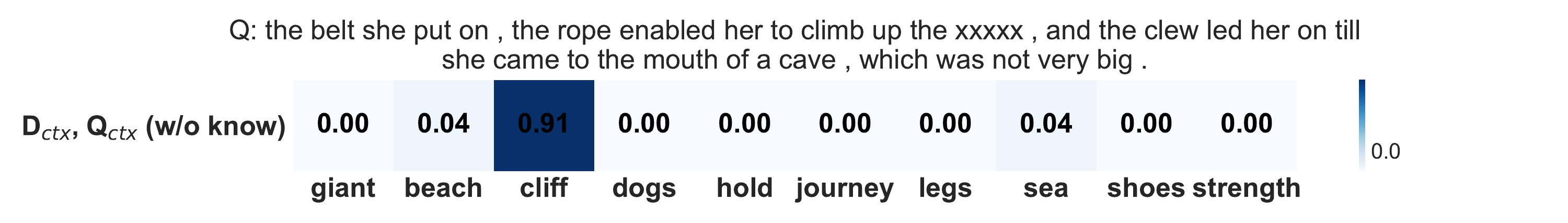}
\endminipage\hfill
\vspace*{-2mm}
\minipage{0.95\textwidth}
  \centering
  \includegraphics[width=\linewidth]{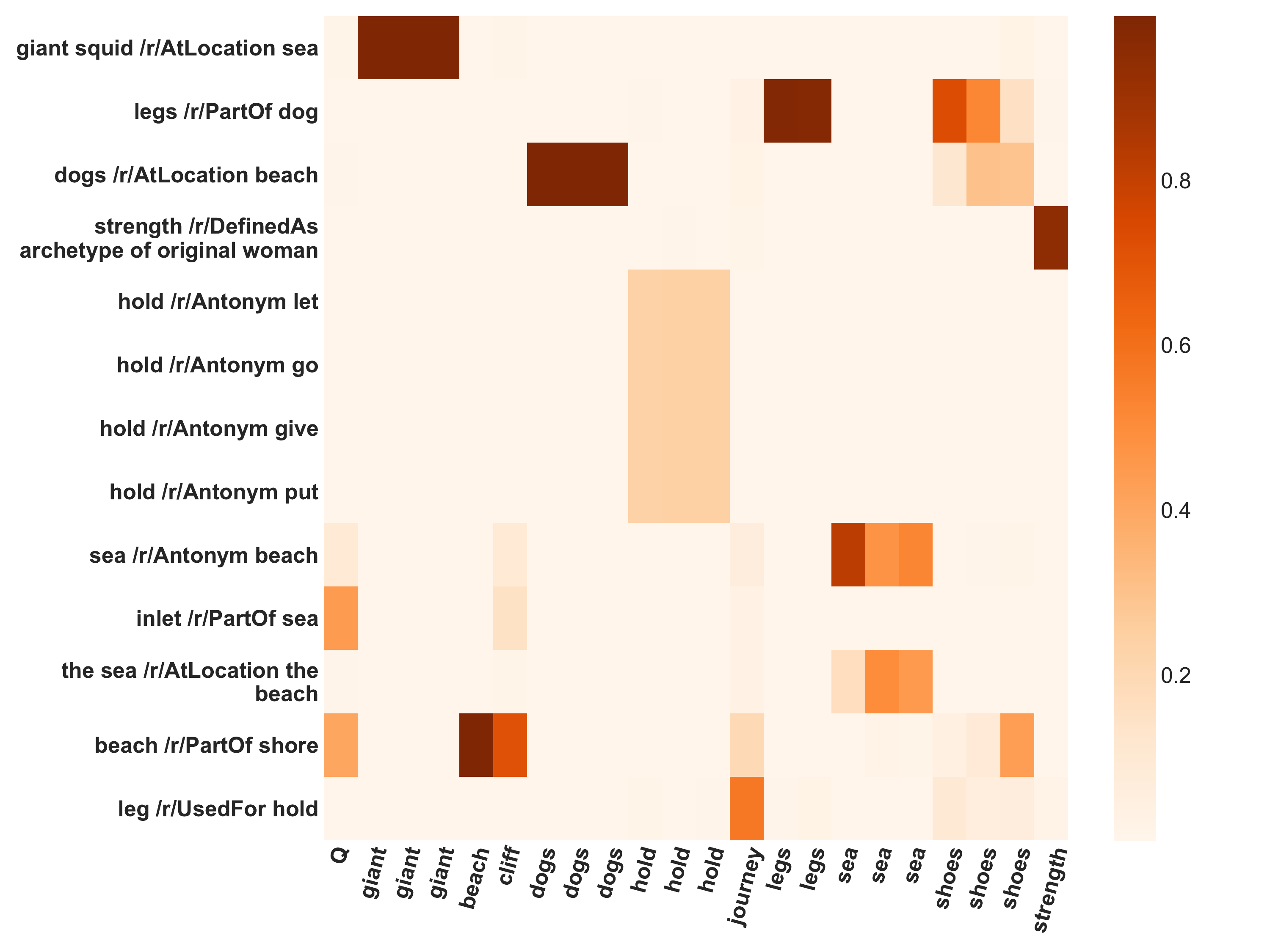}
\endminipage\hfill
\vspace*{-2mm}
\minipage{0.95\textwidth}
  \centering
  \includegraphics[width=\linewidth]{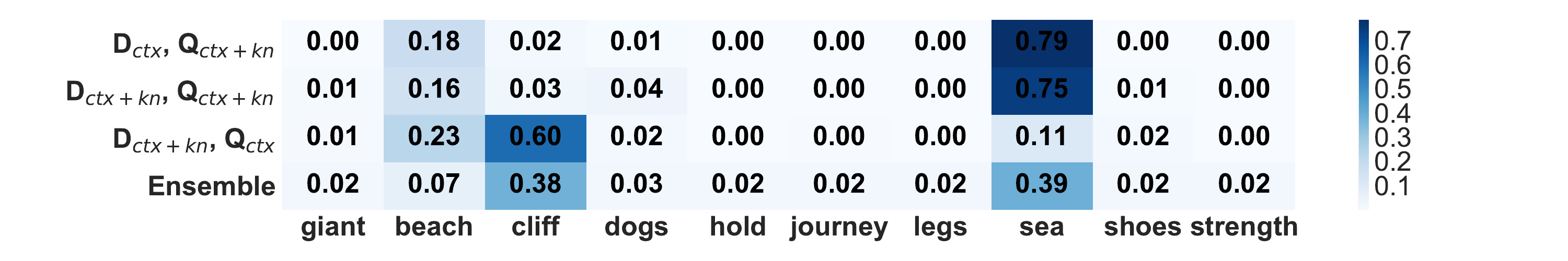}
\endminipage\hfill
\caption{\textbf{Case 5:} Interpreting the components of \emph{KnReader} (\emph{Full model}). Adding retrieved knowledge to $Q$ and $D$ confuses the model and decreases
%helps the model to increase 
the score for the correct answer. Results for top 5 candidates are shown.
(Subj/Obj as key-value memory, 50 facts, \CNSel) (Item \#187)}
\label{fig:item_187}
%\vspace*{-3mm}
\end{figure*}

\paragraph{Case 5}
The example in Figure \ref{fig:item_187} 
\cam{illustrates}
the lack of knowledge about locations. The context of $Q$ talks about \textit{climbing up} and while the text-only module selects the right answer \textit{cliff}, the available knowledge modifies the representation and reverses the answer to \textit{sea} which is \textit{usually} on lower level. Here the association is made with a \textit{cliff} and \textit{sea} by the fact \textit{inlet /r/PartOf sea} and \textit{beach /r/PartOf shore}). That is, the context-only neural representation guesses that the plausible answer is similar to \textit{cliff} (\textit{inlet and shores are usually associated with cliff}). \cam{Again, we are missing knowledge of actions, e.g., that \textit{climbing} is done to move up steep locations such as hills, or cliffs. In future work we plan to experiment with sources that offer more information about events.}
% _cbt_cloze_v15n_f50_cn5nown_tfidf_test

\end{document}